\documentclass[pdflatex,sn-mathphys-num]{sn-jnl}

\usepackage{graphicx}%
\usepackage{multirow}%
\usepackage{amsmath,amssymb,amsfonts}%
\usepackage{amsthm}%
\usepackage{mathrsfs}%
\usepackage[title]{appendix}%
\usepackage{color,xcolor}
\usepackage{textcomp}%
\usepackage{manyfoot}%
\usepackage{booktabs}%
\usepackage{algorithm}%
\usepackage{algorithmicx}%
\usepackage{algpseudocode}%
\usepackage{listings}%
\usepackage{array}
\usepackage{booktabs}
\usepackage{subcaption}
\usepackage{tabularx}
\usepackage{autobreak}
\usepackage{amsmath}
\usepackage{ifthen}
\usepackage[numbers]{natbib}

\theoremstyle{thmstyleone}%
\theoremstyle{thmstyletwo}%

\theoremstyle{thmstylethree}%

\begin{document}

\title[Article Title]{Dynamically Weighted Momentum with Adaptive Step Sizes for Efficient Deep Network Training}


\author*[1]{\fnm{Zhifeng} \sur{Wang}}\email{zfwang@ccnu.edu.cn}
\equalcont{These authors contributed equally to this work.}

\author[1]{\fnm{Longlong} \sur{Li}}\email{llli@mails.ccnu.edu.cn}
\equalcont{These authors contributed equally to this work.}

\author*[2]{\fnm{Chunyan} \sur{Zeng}}\email{cyzeng@hbut.edu.cn}

\affil*[1]{\orgdiv{Faculty of Artificial Intelligence in Education}, \orgname{Central China Normal University}, \orgaddress{\city{Wuhan}, \postcode{430079}, \country{China}}}

\affil[2]{\orgdiv{High-Efficiency Utilization of Solar Energy and Operation Control of Energy Storage System}, \orgname{Hubei University of Technology}, \orgaddress{\city{Wuhan}, \postcode{430068}, \country{China}}}



\abstract{Within the current sphere of deep learning research, despite the extensive application of optimization algorithms such as Stochastic Gradient Descent (SGD) and Adaptive Moment Estimation (Adam), there remains a pronounced inadequacy in their capability to address fluctuations in learning efficiency, meet the demands of complex models, and tackle non-convex optimization issues. These challenges primarily arise from the algorithms’ limitations in handling complex data structures and models, for instance, difficulties in selecting an appropriate learning rate, avoiding local optima, and navigating through high-dimensional spaces. To address these issues, this paper introduces a novel optimization algorithm named DWMGrad. This algorithm, building on the foundations of traditional methods, incorporates a dynamic guidance mechanism reliant on historical data to dynamically update momentum and learning rates. This allows the optimizer to flexibly adjust its reliance on historical information, adapting to various training scenarios. This strategy not only enables the optimizer to better adapt to changing environments and task complexities but also, as validated through extensive experimentation, demonstrates DWMGrad’s ability to achieve faster convergence rates and higher accuracies under a multitude of scenarios. Specifically, our experiments include assessments on classical tasks in computer vision, natural language processing, and audio processing, wherein DWMGrad has demonstrated exceptional classification accuracy and optimization efficiency in CIFAR-10, CIFAR-100 and ImageNet image classification tasks, NLP and graph classification tasks based on the Roberta model. In audio classification tasks, this method has also proven its performance to be comparable to that of classical optimization methods. Furthermore, our algorithm was tested on the Rosenbrock function to further affirm its universality and robustness. Our research not only validates the effectiveness of DWMGrad in traditional deep learning tasks but also showcases its potential in addressing complex optimization challenges. Ultimately, theoretical proofs confirming the convergence properties of the DWMGrad optimizer are provided, further corroborating the effectiveness and feasibility of our approach in surmounting the challenges of deep learning optimization.}




\maketitle

\section{Introduction}\label{sec1}

Deep learning technology has become a mainstream framework, widely applied in various fields such as medical image processing, autonomous driving, educational assessment, real-time translation, and biology \cite{Messeri2024,Wang2025,Shi2026,Wang2025b,Dong2025,Wang2025d,Chen2025b,Wang2024m,Chen2024e,Wang2023v,Ma2023b,Wang2022as,Wang2023j,Liao2024,Wang2024p,Min2019,Wang2023g,Chen2024g}. In medical image processing, deep learning is utilized for segmentation, identification, and classification of medical images, aiding physicians in disease diagnosis and treatment planning  \cite{stokes2020deep,hammernik2023physics}. In the field of autonomous driving, deep learning processes sensor data from vehicles such as cameras, radar, and LiDAR, enabling object detection and recognition, path planning, and environmental perception \cite{grigorescu2020survey,muhammad2020deep,Wang2024m}. In educational assessment, deep learning analyzes student learning behavior data, providing personalized learning support and assessment, thus promoting individualized and intelligent education  \cite{Li2026a,Wang2025e,Li2023i,Wang2024b,Li2023g,Wang2024s,Li2023f,Wang2023j,Dong2023,Wang2023d,Lyu2022,Wang2025f}. In real-time translation, deep learning excels in translation tasks within the domain of natural language processing, facilitating communication and exchange between different languages \cite{zhou2020sign}. In the realm of biology, deep learning is applied in genomics, protein structure prediction, drug discovery, etc., accelerating research and application in biological sciences \cite{jin2021application,saxe2021if}. These applications demonstrate the widespread utilization and significant role of deep learning technology across various domains \cite{Zheng2025,Zeng2025,Chen2025a,Zeng2024g,Chen2025,Zeng2024h,Zheng2024,Zeng2024b,Wang2023f,Zeng2024f,Chen2023b,Zeng2024c,Wang2022t,Zeng2024d,Wang2021m,Zeng2024,Wang2020h,Zeng2024a,Wang2018a,Zeng2023a,Wang2015b,Zeng2023,Zhu2013,Zeng2022a,Wang2011,Zeng2021a,Wang2011a,Zeng2021b,Zeng2020,Zeng2018,Wang2023v,Zeng2025a,Wang2023l,Zeng2024e,Wang2022at,Zeng2023c,Wang2025g,Li2023h,Zeng2022,Wang2022ac,Zeng2022b,Wang2021,Zeng2021c,Tian2018,Zeng2020a,Min2018,Wang2017,Wang2015a}. However, training deep neural networks for non-convex problems is computationally intensive, and finding global minima poses a challenging problem. Therefore, researching optimizers for deep learning networks is of paramount importance, as researchers continuously strive to enhance the performance and convergence speed of deep learning models \cite{verma2023wsagrad,tong2022calibrating}.

In the field of machine learning, optimization algorithms based on gradient computation have long been the primary tools for training neural networks. Among these, Stochastic Gradient Descent (SGD) stands out as a typical gradient-based optimizer, renowned for its simplicity and efficiency, holding significant practical importance across various scientific and engineering domains \cite{robbins1951stochastic,bottou2010large}. The SGD algorithm laid the foundation for subsequent optimization methods based on gradient computation. However, with the advancement of deep learning and computer hardware technology, issues such as slow initial convergence speed and poor performance on sparse data have become increasingly prominent. Particularly in the context of today's large-scale models and datasets in computer science, the limitations of SGD cannot be overlooked \cite{ruder2016overview,dogo2018comparative}. In order to address these challenges, researchers have embarked on exploring directions for optimizing SGD, identifying avenues to enhance its performance through an in-depth examination of its shortcomings. Among these efforts, the introduction of momentum and adaptive learning rate adjustment emerge as two primary directions for further improving the performance of SGD. These improvements aim to expedite the convergence process, enhance the model's generalization capability on complex datasets, thereby addressing challenges encountered in deep learning.

The incorporation of momentum into gradient computation aims to tackle issues like oscillation or slow convergence speed in the SGD algorithm. Polyak introduced the classic momentum gradient descent algorithm in 1964, which utilizes historical gradient information to adjust the update direction of model parameters more effectively. This momentum term ensures updates depend not only on the current gradient but also on past gradients, accelerating movement in the gradient direction and overcoming convergence issues. Variants like Nesterov's Accelerated Gradient (NAG) perform well with significant momentum, yet a high momentum may cause the model to overly rely on past gradients, diminishing generalization capability. Attia et al. demonstrated Nesterov acceleration's instability in initialization and its performance degradation with increasing gradient orders \cite{polyak1964some, nesterov1983method, attia2021algorithmic}. Given SGD's dependency on learning rate and the instability of momentum-based optimization, there's a push towards adaptive algorithms in optimization.

To address these challenges, researchers have proposed numerous adaptive optimization algorithms. In 2011, John Duchi \cite{duchi2011adaptive} et al. introduced the AdaGrad algorithm, which can be seen as the first introduction of adaptive step sizes in the SGD algorithm. This algorithm adjusts the learning rate by accumulating the square of gradients for each parameter, thus performing better when dealing with very sparse gradients or gradients of different scales. A key feature of the AdaGrad algorithm is its ability to alleviate excessive reliance on hyperparameters and meet the varying demands of different dimensions' parameters for different step sizes.  However, AdaGrad has limited ability to generalize to unseen data. In cases where there are considerable dense gradients, the averaging in the denominator leads to small updates, causing AdaGrad to fail. Researchers have proposed replacing the denominator with the square root of an exponential moving average of past gradient squares to address this issue, resulting in optimizers such as RMSprop \cite{hinton2012neural}, Adadelta \cite{zeiler2012adadelta}, Adam \cite{kingma2014adam}, etc. Despite the popularity of the Adam optimizer and its variants due to their faster convergence speed and computational efficiency, there are still theoretical concerns about their instability and slow convergence \cite{reddi2019convergence,wilson2017marginal}. In order to achieve a balance between momentum adjustment and adaptive learning rate adjustment, in this paper, we focus on historical information related to the model optimization process. We believe that considering relevant historical information within an appropriate time frame can accelerate the convergence speed of network models and help approximate the optimal solution. We propose a method to dynamically adjust the size of the window to determine the range of relevant historical information. Based on this concept, we develop rules for updating gradients and momentum. This strategy leads to the development of the DWMGrad Optimizer, an optimizer that dynamically adjusts momentum and learning rate based on historical information. Experiments conducted on various machine learning models and datasets demonstrate that our optimizer outperforms other algorithms in terms of convergence. Additionally, we provide theoretical analysis and convergence proofs of DWMGrad in the context of convex optimization. In summary, our contributions can be summarized as follows:

\begin{itemize}
	\item We introduce the DWMGrad optimization framework, which, thanks to a mechanism for precisely adjusting key parameters in dynamic environments, allows our method to adapt to various optimization scenarios and demands. In addition, we provide a theoretical analysis proving the convergence properties for convex optimization problems, alongside a mathematical model for the control of learning rates and momentum.
	\item This approach utilizes dynamic variable Windows to define the temporal dimension of the historical information related to model optimization and regulates the dynamic range of historical information to enhance the algorithm’s generalization ability.
	\item We establish an update rule algorithm based on historical gradients and momenta. This kind of strategy, driven by historical information, enables accelerated algorithm convergence through learning rate and momentum adjustments.
	\item The DWMGrad optimizer has been empirically validated as highly effective for various deep learning tasks. In computer vision, it has achieved outstanding classification accuracy and efficiency. Similarly, it excels in natural language processing and graph classification tasks, showcasing superior performance. In audio classification, DWMGrad competes well with traditional optimization methods, highlighting its broad utility. Additionally, tests using the Rosenbrock function have confirmed DWMGrad's universality and robustness, proving its capability to tackle complex optimization challenges effectively.
\end{itemize}

The remaining parts of this paper are organized as follows: Section \ref{sec:2} reviews related work on current optimizers. Section \ref{sec:3} elaborates on the proposed DWMGrad optimizer. Section \ref{sec:4} provides a brief overview of the experimental details and compares the performance of the proposed method with several benchmark models and SOTA approaches. Section \ref{sec:5} summarizes the entire paper and discusses future directions for optimizer.

\section{Related Work}
\label{sec:2}
This section begins by formally defining the optimization problem for neural networks. It then reviews existing approaches from three perspectives: optimization algorithms based on gradients, those utilizing momentum, and algorithms that employ adaptive learning rates. Here, we establish some notational conventions to facilitate the description of various optimization algorithms in the subsequent sections, which are described in detail in Table \ref{t:1}.

\begin{table}[t]
	\centering
	\caption{Main symbols and descriptions.}
	\begin{tabular}{cp{0.8\linewidth}}
		\hline
		Symbol & Description \\
		\hline
		$f(\theta)$ & Stochastic objective function with parameters $\theta$. \\
		$g_t$ & Gradient at moment $t$. \\
		$\alpha$ & Learning rate. \\
		$\beta$ & Loss difference accumulator. \\
		$\gamma$ & Momentum. \\
		$\theta$ & Parameter vector. \\
		$\omega$ & Window size for historical series data. \\
		$t$ & Time step. \\
		$\delta$ & Maximum window size. \\
		$\epsilon$ & An auxiliary quantity that prevents the value from being calculated as 0. \\
		\hline
	\end{tabular}
	\label{t:1}
\end{table}

\subsection{Neural Network Optimization Problem}

The optimizer serves an indispensable role in the domain of machine learning by minimizing the loss on the training set to guide the learning process of the model parameters. Gradient-based optimization methods entail the use of iterative processes to minimize the parameter vector of a target function $F(\theta)$ \cite{mcrae2021memory,wilson2017marginal}. Formally, one considers a bounded scalar function $f$: $R^d\to R$, where $R$ denotes the realm of real numbers, $d$ signifies the dimensionality of the parameters, and $R^d$ represents the Euclidean space of dimension d. The optimization challenge is to solve for $\min_{\theta\in R^d}f(\theta)$. The optimal parameter vector for the aforementioned objective function, denoted by $\theta^*$, thus the optimization goal can be articulated as:
\begin{equation}
	\theta^*=arg\min_{\theta\in{R}^d}f(\theta)
\end{equation}

With the progression of time steps through iterations, the optimizer iteratively updates the parameter vector toward convergence with the optimal solution, expressed as:
\begin{equation}
	\lim_{t\to\infty}\parallel\theta_t-\theta^*\parallel=0
\end{equation}

\subsection{Neural Network Optimization Algorithms}
The preceding description provided an overview of the optimization framework based on gradient computation. After that,  we will introduce the commonly used optimization algorithms as well as the latest improved algorithms.

\subsubsection{Gradient Descent Based Algorithms}
\textbf{Batch Gradient Descent (BGD)} \cite{Ahmadianfar2020a} is a traditional gradient optimization algorithm widely used to train machine learning models, especially neural networks. It fundamentally computes the model parameters' gradients using the entire training dataset to update these parameters. BGD stabilizes the update direction by employing the full dataset for gradient calculations in each iteration. However, this method requires processing the entire dataset for every update, which can significantly slow down computation and prolong training times with large datasets. Additionally, it necessitates loading the complete dataset into memory, which can lead to excessive memory consumption.

\textbf{Stochastic Gradient Descent (SGD)} \cite{robbins1951stochastic} is an improvement upon the traditional gradient descent algorithm designed to efficiently handle large-scale datasets. As one of the most widely used optimization algorithms, SGD holds significant importance in the field of deep learning. Its update rule is represented by the following formula:
\begin{equation}
	\theta_{t+1}=\theta_t-\alpha\nabla L(\theta_t;x_i,y_i)
\end{equation}

The stochastic nature of SGD arises from updating parameters after computing the gradient for each data point. This randomness may introduce noise into the optimization process, but it also allows SGD to escape local minima and find better solutions in high-dimensional spaces. Due to the stochastic nature that can introduce noise into the optimization process, resulting in noisy convergence behavior, SGD has been accompanied by various variants, such as mini-batch SGD and momentum-based SGD algorithms, to address these challenges. 

\textbf{Mini-batch SGD} \cite{woodworth2020local} algorithm divides the dataset into mini-batches and computes gradients for each mini-batch to update model parameters. This approach combines the advantages of batch gradient descent and stochastic gradient descent, enhancing the stability of parameter updates while maintaining a certain level of randomness. This is beneficial for preventing the algorithm from getting trapped in local minima.

\subsubsection{Momentum Based Algorithms}

\textbf{Momentum-SGD (MSGD)} introduces a momentum term during parameter updates, which mimics the concept of inertia in physics. In the parameter update process, it not only considers the current gradient but also takes into account previous velocity and direction \cite{ruder2016overview}. This operation reduces fluctuations in parameter updates in the horizontal direction, leading to smoother convergence, accelerated convergence speed, and avoidance of the potential of the loss function getting trapped in elongated valleys. The computational formula for parameter updates is as follows:
\begin{equation}
	\left\{
	\begin{aligned}
		& v_{t+1} = \gamma v_t + \alpha \nabla L(\theta_t) \\
		& \theta_{t+1} = \theta_t - v_{t+1}
	\end{aligned}
	\right.
\end{equation}

The introduction of this momentum term means that the update direction is influenced not only by the current gradient but also by the historical gradient directions. This helps reduce fluctuations in parameter updates in the vertical direction and accelerates updates in the horizontal direction. Consequently, the algorithm is aided in escaping local minima and converging more rapidly to the global optimum.

\textbf{Nesterov's Accelerated Gradient (NAG)} MSGD considers only the weighted average of past gradients \cite{nesterov1983method}. Yurii Nesterov proposed the Nesterov's accelerated gradient (NAG) algorithm in 1983, which provides a more refined estimation of the next parameter position. This approach helps avoid the "oscillation" phenomenon that may occur when gradients are large, thereby accelerating convergence and improving training efficiency. The computational formula for parameter updates is as follows:
\begin{equation}
	\left\{
	\begin{aligned}
		& v_{t+1} = \gamma v_t + \alpha \nabla L(\theta_t - \gamma v_t) \\
		& \theta_{t+1} = \theta_t - v_{t+1}
	\end{aligned}
	\right.
\end{equation}

In NAG, the gradient is estimated at the expected position $\theta_t-\gamma v_t$ rather than the current position $\theta_t$, making the update direction more accurate by considering the influence of the momentum term on the next position. Moreover, compared to standard momentum SGD, it is more effective in avoiding overshooting of parameter updates. Another method that has emerged as an improvement to SGD is adaptive adjustment of the step size, leading to the development of many excellent optimizers.

\subsubsection{Adaptive Learning Rate Based Algorithms}

\textbf{The Adaptive Gradient Algorithm (AdaGrad)} is a representative adaptive optimizer algorithm that dynamically adjusts the learning rate based on the historical gradient information of each parameter to accelerate the convergence process of the model \cite{duchi2011adaptive}. Specifically, it reduces the learning rate for parameters with large sparsity and increases it for parameters with small sparsity. The update rule for AdaGrad can be expressed as follows:
\begin{equation}
	\left.\left\{\begin{aligned}G_{t+1}&=G_t+(\nabla L(\theta_t))^2\\\theta_{t+1}&=\theta_t-\frac\alpha{\sqrt{G_{t+1}+\epsilon}}\cdot\nabla L(\theta_t)\end{aligned}\right.\right.
\end{equation}

Here, $G_{t+1}$ represents the accumulated squared gradients at time step $t+1$, which is used for adjusting the learning rate. Because AdaGrad adaptively adjusts the learning rates, allowing each parameter to have its own learning rate, it is suitable for non-convex optimization problems and sparse datasets. Conversely, due to the learning rates potentially becoming too small over time, leading to premature termination of the training process, researchers have proposed new algorithms to overcome these difficulties. RMSprop is one of the most representative algorithms among them.

\textbf{The Root Mean Square Propagation (RMSprop)} algorithm aims to address the issue of premature learning rate decay in AdaGrad \cite{hinton2012neural}. It utilizes exponentially weighted moving averages to calculate the exponentially decaying average of squared gradients, thereby alleviating the problem of premature learning rate decay in the AdaGrad algorithm. The parameter update rule for the RMSprop algorithm is as follows:
\begin{equation}
	\left.\left\{\begin{aligned}&E[g^2]_{t+1}=\beta E[g^2]_t+(1-\beta)(\nabla L(\theta_t))^2\\&\theta_{t+1}=\theta_t-\frac\alpha{\sqrt{E[g^2]_{t+1}+\epsilon}}\cdot\nabla L(\theta_t)\end{aligned}\right.\right.
\end{equation}

In this context, $[g^2]_t$ denotes the exponentially weighted moving average of the squared gradients at time step $t$. $\beta$ represents the decay rate of the exponential moving average, typically ranging between $0$ and $1$. These strategies in RMSprop contribute to a smoother computation of the learning rate. By introducing the decay rate, RMSprop is capable of better adapting to gradient changes, slowing down the descent rate of the learning rate, thus rendering the training process more stable. Therefore, RMSprop is more suitable for handling non-stationary objective functions or situations with significant gradient noise.

\textbf{Adaptive Moment Estimation (Adam)} is an optimization algorithm widely used in the field of deep learning, which combines momentum adjustment and adaptive learning rate adjustment \cite{kingma2014adam}. Its core idea is to combine the advantages of momentum and adaptive learning rate, allowing for stable updates while using different learning rates for different parameters. It maintains two first-moment estimates (i.e., the mean of gradients) and second-moment estimates (i.e., the exponentially weighted moving average of gradients' squared values), and then uses them to adjust the learning rates for each parameter. The update rule of Adam is as follows:
\begin{equation}
	\begin{cases}\begin{aligned}m_{t+1}&=\beta_1m_t+(1-\beta_1)\nabla L(\theta_t)\\v_{t+1}&=\beta_2v_t+(1-\beta_2)(\nabla L(\theta_t))^2\\\widehat{m}_{t+1}&=\frac{m_{t+1}}{1-\beta_1^{t+1}}\\\hat{v}_{t+1}&=\frac{v_{t+1}}{1-\beta_2^{t+1}}\\\theta_{t+1}&=\theta_t-\frac\alpha{\sqrt{\widehat{v}_{t+1}}+\epsilon}\cdot\widehat{m}_{t+1}\end{aligned}&\end{cases}
\end{equation}

Here, $m_t$ represents the first-moment estimate at time step $t$, which is the exponentially weighted moving average of gradients. $v_t$ represents the second-moment estimate at time step $t$, which is the exponentially weighted moving average of the squared gradients. $\beta_1$ and $\beta_2$ are momentum parameters used to control the update rates of the first and second moments, respectively. $\hat{m}_{t+1}$ and $\hat{v}_{t+1}$ are bias-corrected estimates of the first and second moments, used to address the estimation bias issue in the initial stages. The Adam algorithm adjusts the learning rates adaptively across different parameters using the momentum term $m_t$ and the second-moment term $v_t$. Additionally, due to the use of exponential moving averages, the Adam algorithm provides a good estimate of the trend of gradients, leading to more stable parameter updates. Therefore, the Adam algorithm is typically effective in optimizing loss functions and accelerating model convergence. With the expansion of the field of artificial intelligence and the emergence of various tasks, variants of the Adam optimizer have been proposed, such as AdamW, NAdam, RAdam, etc.

\textbf{AdamW} is an improvement upon the Adam algorithm. In the standard Adam algorithm, weight decay is typically directly added to the parameter update step, wherein the regularization term of weight decay is added to the gradient during gradient computation, followed by parameter update \cite{loshchilov2017decoupled}. However, this approach may lead to incorrect handling of weight decay during training, especially when the learning rate is small. AdamW separates weight decay from the parameter update step and applies it independently to the parameters. Specifically, in AdamW, weight decay is applied before parameter updates. This is done to alleviate the impact of weight decay on the learning rate, thereby enhancing the stability and generalization performance of the algorithm.

\textbf{Rectified Adam (RAdam)} aims to address the performance degradation issue that may arise from an excessively high learning rate during the initial stages of training \cite{fall2016risk}. The RAdam algorithm introduces a mechanism for adaptive learning rate correction, wherein the learning rate is dynamically adjusted based on the magnitude of gradients during learning rate computation. Specifically, RAdam considers a dynamic upper limit for the learning rate calculation, ensuring that the learning rate does not become excessively high during the initial stages of training.

\textbf{Nesterov-accelerated Adaptive Moment Estimation (NAdam)} combines Nesterov momentum with the Adam algorithm \cite{ruder2016overview}. NAdam utilizes Nesterov momentum in the computation of exponentially weighted moving averages of gradients, thereby ensuring more accurate parameter updates. Compared to the Adam algorithm, NAdam performs better on some optimization problems, especially those involving complex loss functions.

\textbf{Summry}: This section provides descriptions of some classic and popular optimization algorithms, each of which plays a unique role in a particular domain. These improved algorithms typically offer better performance and convergence speed in practice, especially when training deep neural networks. The choice of algorithm suitable for a specific task depends on the nature of the problem and the factors that need to be balanced, such as algorithm stability, generalization ability, and computational resources.

\section{Proposed Method}
\label{sec:3}

This section begins with a comprehensive introduction of the DWMGrad Optimizer framework, detailing the update rules for each component. Subsequently, it delves into a theoretical analysis of DWMGrad's convergence and computational complexity.
\subsection{Framework of DWMGrad Optimizer}

\begin{figure}[ht]
	\centering
	\includegraphics[width=\textwidth]{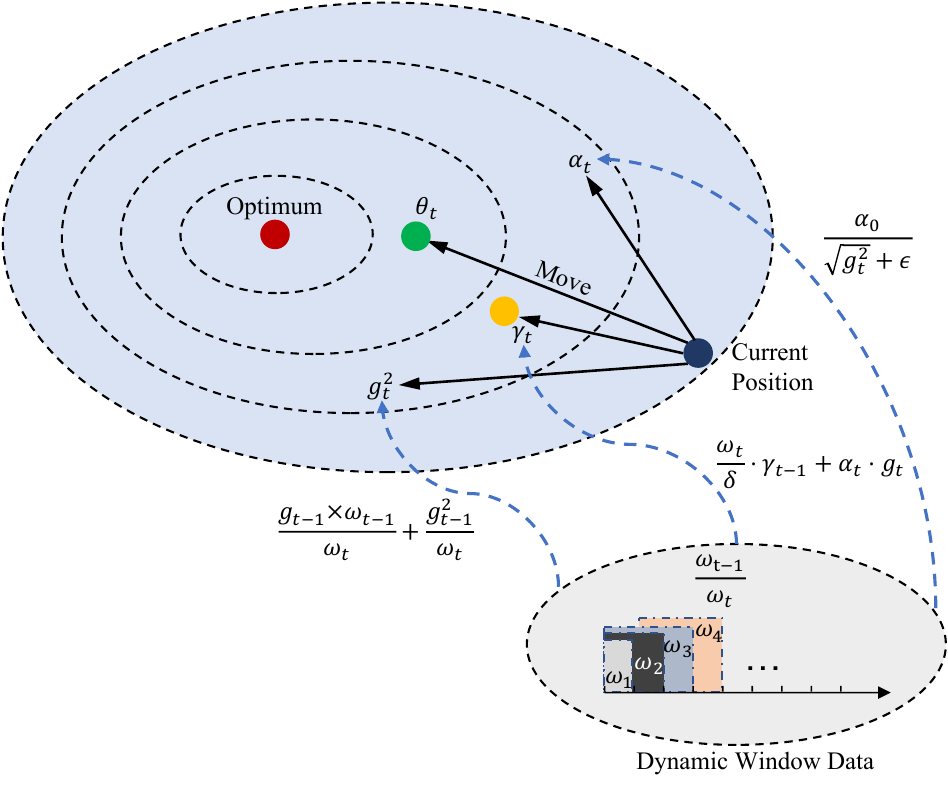}
	\caption{Diagram of the DWMGrad optimization framework.}
	\label{f:2}
\end{figure}

In this section, we investigate a gradient descent algorithm that is enhanced by a dynamic historical information framework to optimize the adjustment of momentum and learning rates. The limitation of current algorithms lies in their static consideration of historical data, which influences parameter adjustment. To address this, our method incorporates a dynamic mechanism that adjusts the span of historical data considered during optimization. Specifically, our optimizer dynamically adjusts the historical information window, which influences parameter updates. This historical information pertains to the data collected in the parameter space throughout the gradient descent process. Initially, we establish a window and define its initial and maximum dimensions, analogous to the $\beta_{1}$ and $\beta_{2}$ parameters in the Adam optimizer. Our approach offers the advantage of a flexible, dynamically adaptive window, which surpasses the rigidity of the fixed parameter span in Adam. The details of our optimizer are presented in Algorithm \ref{alg1}. Fig. \ref{f:2} depicts the behavior of the individual parameters in the algorithm.

\begin{algorithm}
\caption{DWMGrad Optimizer}
\label{alg1}
\begin{algorithmic}[1] 
\State Initialize $\theta_0$ randomly; $\omega \leftarrow 5$; $\gamma \leftarrow 0.9$; $\delta \leftarrow 10$; $\beta \leftarrow 0$; $\alpha_0 \leftarrow 0.001$; $\epsilon \leftarrow 10^{-8}$
\While{$\theta_t$ not converged}
    \State Sample batch of meta-training tasks $\mathcal{T}_i \sim p(\mathcal{T})$
    \State $g_t = \nabla f(\theta_{t-1})$
    \State $\beta \leftarrow \beta + \text{loss from } \mathcal{T}_i$
    \If{$\beta > 0$}
    \Comment{Adjust the $\omega$ according to the $\beta$}
        \State $\omega_t \leftarrow \min(\omega_{t-1} + 1, \delta)$
    \Else
        \State $\omega_t \leftarrow \max(\omega_{t-1} - 1, 1)$
    \EndIf
    \State $g_t^2 \leftarrow \frac{g_{t-1} \times \omega_{t-1}}{\omega_t} + \frac{g_{t-1}^2}{\omega_t}$ \Comment{Weighted gradient square}
    \State $\alpha_t \leftarrow \frac{\alpha_0}{\sqrt{g_t^2} + \epsilon}$ \Comment{Compute adaptive learning rate}
    \State $\gamma_t \leftarrow \frac{\omega_t}{\delta} \cdot \gamma_{t-1} + \alpha_t \cdot g_t$ \Comment{Update velocity with adaptive learning rate}
    \State $\theta_t \leftarrow \theta_{t-1} - \gamma_t$  \Comment{Update parameters}
\EndWhile
\State \Return $\theta_t$
\end{algorithmic}
\end{algorithm}

\subsection{Update Rules of DWMGrad Optimizer}

In this work, a fundamental characteristic of the DWMGrad optimizer lies in its capability to dynamically leverage historical information to refine and enhance current parameter updates. The initialization of the DWMGrad parameters is specified as follows: $\theta_0$ is initialized randomly, $\omega \leftarrow 5$, $\gamma \leftarrow 0.9$, $\delta \leftarrow 10$, $\beta \leftarrow 0$, $\alpha_0 \leftarrow 0.001$, and $\epsilon \leftarrow 10^{-8}$.

\textbf{The update of the squared gradient ${g_{t}}^{2}$}: this corresponds to the first component of the core update rules. As shown in Eq. \ref{eq:9}, computes a weighted average of the squared gradient ${g_{t}}^{2}$, where the weights are contingent on the window size $\omega$ at the current and previous time steps.  

\begin{equation}
	g_t^2 \leftarrow \frac{g_{t-1} \times \omega_{t-1}}{\omega_t} + \frac{g_{t-1}^2}{\omega_t}
	\label{eq:9}
\end{equation}

Specifically, the term $\frac{g_t \cdot \omega_{t-1}}{\omega_t}$ in the formula reflects the impact of the previous time step's window size on the current gradient, while $\frac{g_t^2}{\omega_t}$ represents the normalization of the current squared gradient. This formulation encapsulates the varying influence of different window sizes on the sensitivity of the algorithm during distinct phases of parameter updates. It assists in moderating the rate of learning when encountering noise or irregular loss surfaces. For descriptive purposes, we refer to $\omega_t$ in this update step as the dynamic adjustment factor. Overall, this factor is employed to more effectively modulate the influence of historical weights and adaptively adjusts as parameter iterations progress, thereby informing parameter updates.

\textit{Motivation}:  
The squared gradient update aims to stabilize optimization by balancing historical and current gradient information. Using an exponential decay mechanism controlled by the window size $\omega_t$, it dynamically adjusts the influence of past gradients. This helps reduce gradient noise and variance. Specifically, the normalized squared gradient term ($\frac{g_{t-1}^2}{\omega_t}$) smooths sudden changes in gradient magnitude and corrects directional deviations in updates. The window parameter $\omega_t$ adjusts over time: in early training, it favors quick response to new gradients, while in later stages, it promotes smoother, more stable updates.  

\textbf{The update of learning rate $\alpha_t$}: The update of $\alpha_t$ embodies the concept of adaptive learning rate. It represents the adjustment of the learning rate by the normalization of the current gradient with respect to the square of the gradient, as defined in Eq. \ref{eq:10}. The term $\sqrt{g_t^2}$ ensures that the magnitude of the gradient does not adversely affect the update of the learning rate, and $\epsilon$ is a small constant added to stabilize the numerical computation.

\begin{equation}
	\alpha_t \leftarrow\frac {\alpha_0} {\sqrt{g_t^2}+\epsilon}
	\label{eq:10}
\end{equation}

\textit{Motivation}:  
The learning rate update adaptively adjusts the step size based on the gradient magnitude to ensure stable and efficient optimization. By normalizing the learning rate with $\sqrt{g_t^2}$, the method reduces the step size in steep regions to avoid overshooting and increases it in flat regions to speed up convergence. A small constant $\epsilon$ is added to maintain numerical stability. This adaptive strategy helps the algorithm balance exploration and exploitation, handle varying gradient scales, and achieve better convergence across different tasks. 

\textbf{The update of momentum $\gamma_t$}: The momentum method is inspired by the concept of momentum in physics, assisting parameters to overcome local minima, reduce oscillations, and hasten convergence during the optimization process. As defined in Eq. \ref{eq:11}, the term $\frac{\omega_t}{\delta} \cdot \gamma_{t-1}$ represents the adjustment of the previous time step's momentum in proportion to the current window size $\omega_t$ relative to the maximum window size $\delta$.

\begin{equation}
	\gamma_t\leftarrow\frac{{\omega_t}}\delta\cdot\gamma_{t-1}+\alpha_t\cdot g_t
	\label{eq:11}
\end{equation}

\textit{Motivation}:  
The momentum update balances past momentum and current gradients to improve stability during optimization. By scaling the previous momentum $\gamma_{t-1}$ with the ratio $\frac{\omega_t}{\delta}$, the algorithm adaptively controls how much historical information is retained. A larger window size $\omega_t$ preserves more momentum, increasing inertia to help escape local minima, while a smaller $\omega_t$ emphasizes recent gradients for quicker adaptation. This mechanism mimics physical inertia: consistent gradients build momentum to speed up convergence in flat regions, while reducing oscillations in sharp areas. Paired with the adaptive learning rate $\alpha_t$, the momentum term $\gamma_t$ provides a coordinated adjustment of both direction and step size, enhancing convergence across different landscapes.

\textbf{The update of the parameter $\theta$}: The update of parameter $\theta$ follows the standard gradient descent rule, as described in Eq. \ref{eq:12}. The current update $\theta_t$ is achieved by subtracting the momentum term $\gamma_t$ from the parameter value at the previous time step $\theta_{t-1}$.  
\begin{equation}  
	\theta_{t}\leftarrow\theta_{t-1}-\gamma_{t}  
	\label{eq:12}  
\end{equation}  

The term $\gamma_t$ encompasses information about the gradient direction and historical gradients. This adjustment serves to smooth the parameter update process, preventing oscillations caused by overly large parameter updates while also accelerating the training process.

\textit{Motivation}:  
The parameter update combines momentum smoothing and adaptive learning rate scaling in a unified framework. Instead of directly using the raw gradient $g_t$, it employs the momentum term $\gamma_t$, which integrates both historical trends ($\frac{\omega_t}{\delta} \cdot \gamma_{t-1}$) and adaptively scaled updates ($\alpha_t \cdot g_t$). This dual mechanism helps maintain stability in flat regions through consistent momentum, while reducing oscillations in sharp regions via gradient normalization. By implicitly linking learning rate and momentum dynamics, the update rule reduces the need for manual hyperparameter tuning and achieves faster, more stable convergence across various optimization tasks.   

\textbf{Overall Method Summary:}  
The proposed algorithm is a gradient descent framework that synergizes momentum-driven direction smoothing and adaptive learning rate scaling, governed by dynamic historical gradient information. The balance between historical and current gradient influences is regulated by two key hyperparameters: the \textbf{dynamic window size} $\omega_t$, which acts as a time-varying decay factor controlling the trade-off between historical momentum retention (via $\frac{\omega_t}{\delta}\gamma_{t-1}$) and current gradient contribution (via $\alpha_t g_t$), and the \textbf{maximum window size} $\delta$, which normalizes the historical momentum term $\gamma_{t-1}$ to the range $[0,1]$ through the ratio $\frac{\omega_t}{\delta}$, thereby preventing unbounded growth of historical influence. 
The monotonic increase of $\omega_t$ during training ($\omega_t = \omega_{t-1} + \Delta$) shifts the focus from exploration (early stage) to exploitation (late stage), enabling phase-specific adaptation. In flat regions, increased $\omega_t$ enhances historical momentum to maintain update consistency while the learning rate $\alpha_t$ adaptively grows via Eq.~\ref{eq:10} to accelerate traversal; in steep regions, decreased $\omega_t$ prioritizes current gradients and reduces $\alpha_t$ via Eq.~\ref{eq:10} to prevent overshooting.
By integrating momentum smoothing (Eq.~\ref{eq:11}) and adaptive learning rate normalization (Eq.~\ref{eq:10}), the method achieves automatic coordination between step size ($\alpha_t$) and direction ($\gamma_t$) adjustments, eliminating manual hyperparameter coupling while ensuring robustness to gradient scale variations through the $\frac{1}{\sqrt{g_t^2}+\epsilon}$ normalization in Eq.~\ref{eq:10}. 
Unlike traditional methods requiring separate tuning of learning rates and momentum coefficients or fixed exponential decay rates (e.g., Adam), our approach achieves fully automated adaptation through the $\omega_t/\delta$-controlled dynamic windowing mechanism, resulting in faster convergence and improved stability across varying optimization landscapes.

\subsection{Covergence Analysis of DWMGrad}

In the context of dynamical systems theory, the potential energy function serves as a tool for proving the stability of the system. In optimization theory, it can be utilized to demonstrate the convergence of a sequence of parameters. We define a potential function $U(\theta)$, The optimization problem we need to solve is:

\begin{equation}\min_{\theta\in\mathbb{R}}U(\theta)\end{equation}

Next, we have the following constraint on the convergence of DWMGrad:

Theorem 1.\label{t2.1} Assume that the gradient of the objective function $f$ is bounded, i.e., $\|\nabla f_t (\theta)\|_2 \leq G$ for all parameters $\theta \in \mathbb{R}^d$, and that any parameter $\theta_t$ obtained by DWMGrad Optimizer is bounded. Thus, for any time step $t$, the square of the gradient $g_t^2$ has an upper bound $g_t^2 \leq \frac{2G^2}{\omega_{\text{min}}}$, where we assume a bound $M$ such that, based on the properties of $g_t^2$, we obtain $|\gamma_t| \leq \gamma_{\text{max}}$. This upper bound ensures that the gradient does not grow indefinitely with parameter updates. Consequently, we establish that there exists an upper bound for the parameter sequence:
\begin{equation}
	\{\frac{t}{t+1}\sum_{i=0}^{t}\theta_{i}\}_{t=0}^{\infty}\leq\frac{t+1}{t}\parallel\theta_{0}\parallel+\gamma_{max}
\end{equation}

The aforementioned theorem indicates that, when DWMGrad Optimizer is used to optimize a convex function, the generated sequence of parameters is bounded. This result is crucial for ensuring the numerical stability of the optimization algorithm during the iteration process, particularly in the fields of machine learning and deep learning, as it guarantees that the parameters will not diverge due to excessively large updates.

\textit{Step1: Convex Optimization Algorithms Dwmgrad Optimizer Proof of Boundedness}

\textit{Step1.1: Preliminary Lemmas}

\textbf{Lemma 1.} Let \( f: \mathbb{R}^d \rightarrow \mathbb{R} \) be a function such that for all \( x, y \in \mathbb{R}^d \) and any \( \lambda \in [0, 1] \), the following inequality holds:
\begin{equation}
	\lambda f(x) + (1-\lambda) f(y) \geq f(\lambda x + (1-\lambda) y)
\end{equation}

\textbf{Lemma 2.}
Based on convex optimization theory, consider a function \( f: \mathbb{R}^d \rightarrow \mathbb{R} \). It follows that for all \( x, y \in \mathbb{R}^d \):
\begin{equation}
	f(y) \geq f(x) + \nabla f(x)^T (y - x)
\end{equation}

\textit{Step1.2: Gradient Boundedness}

The aforementioned lemmas facilitate the proof of the convergence properties of our optimizer. Given that the gradient of the objective function \( f(\theta) \) is bounded throughout the entire parameter space, there exists a constant \( G > 0 \) such that for all \( \theta \), the \( \ell_2 \)-norm of the gradient satisfies:
\begin{equation}
	g^t = \|\nabla f(\theta)\|_2 \leq G
\end{equation}

Based on the above, we can establish a bound for \( g_t^2 \). Therefore, for any time step, there exists:
\begin{equation}
	g_t^2 \leq \frac{G \cdot G}{\omega_t} + \frac{G^2}{\omega_t} = \frac{2G^2}{\omega_t}
\end{equation}

Consequently, there exists \( \omega_{\min} \) such that \( 0 < \omega_{\min} < \omega_t \) holds for all time steps \( t \), implying that \( g_t^2 \) is upper bounded by \( \frac{2G^2}{\omega_{\min}} \).

\textit{Step1.3: Momentum Boundedness}

When \( t = 0 \), the parameter \( \gamma \) is initialized to \( 0 \) or another value, indicating that the initial value of momentum is bounded.

Assume that at \( t = k \), for some positive constant \( B_{k-1} \), the momentum \( \gamma_{k-1} \) is bounded, i.e., \( |\gamma_{k-1}| \leq B_{k-1} \).

At the time step \( t = k + 1 \), according to the momentum update rule, we have:
\begin{equation}
	\gamma_k = \frac{\omega_k}{\delta} \cdot \gamma_{k-1} + \frac{\alpha \cdot g_k}{\sqrt{g_k^2} + \epsilon}
\end{equation}

Since we have already established that the square of the gradient \( g_t^2 \) is bounded, let us assume a bound \( M \) such that for all time steps \( t \), we have:
\begin{equation}
	\frac{\alpha \cdot g_k}{\sqrt{g_k^2} + \epsilon} \leq \frac{\alpha \cdot M}{\sqrt{\frac{2M^2}{\omega_{\min}}} + \epsilon} \leq M
\end{equation}

Given the algorithmic parameter constraint \( \frac{\omega_t}{\delta} < 1 \), we can rewrite \( g_k \) as follows:
\begin{equation}
	\gamma_k \leq \frac{\omega_k}{\delta} \cdot B_{k-1} + G \leq B_{k-1} + M
\end{equation}

Therefore, it can be deduced that the momentum \( \gamma_k \) is bounded for all time steps. This can be described as:
\begin{equation}
	|\gamma_t| \leq \gamma_{\max}, \forall t
\end{equation}

\textit{Step1.4: Parameter Sequence Boundedness}

In order to prove that the sequence of parameters \( \theta_t \) is bounded, we need to show that the following expression is bounded:
\begin{equation}
	\left\{\frac{t}{t+1}\sum_{i=0}^t\theta_i\right\}_{t=0}^\infty
\end{equation}

Continuing the derivation from the update rule, we obtain (Eq .\ref{eq:25c}):

\begin{equation}
	\label{eq:25a}
	\theta_t = \theta_{t-1} - \gamma_t = \theta_0 - \sum_{i=1}^t \gamma_i
\end{equation}

\begin{equation}
	\label{eq:25b}
	\sum_{i=1}^t \theta_i = \theta_0 + \theta_1 + \theta_2 + \cdots + \theta_t
\end{equation}

\begin{equation}
	\label{eq:25c}
	\sum_{i=1}^t \theta_i = \theta_0 + (\theta_0 - \gamma_1) + (\theta_0 - \gamma_1 - \gamma_2) + \cdots + \left(\theta_0 - \sum_{i=1}^t \gamma_i\right) = (t+1) \theta_0 - \left(\sum_{i=1}^t i \gamma_i\right)^\omega
\end{equation}

Since \( g^t \) is bounded, we have \( |\gamma_t| \leq \gamma_{\max} \). It is necessary to bound the term \( \sum_{i=1}^t i \gamma_i \).
\begin{equation}
	\left|\sum_{i=0}^t i \gamma_i\right|  \leq \sum_{i=0}^t i \gamma_{\max} = \gamma_{\max} \sum_{i=0}^t i = \gamma_{\max} \cdot \frac{t(t+1)}{2}
\end{equation}

\textit{Step1.5: Final Simplification and Conclusion}

Hence, the absolute value of the cumulative sum of the parameter sequence can be bounded by:
\begin{equation}
	\left|\sum_{i=0}^t \theta_i\right| \leq (t+1) \|\theta_0\| + \gamma_{\max} \cdot \frac{t(t+1)}{2}
\end{equation}

It is evident that this is a quadratic function in terms of \( t \), which grows unbounded as \( t \) increases. Consequently, the cumulative sum of the parameters \( \sum_{t=0}^\infty \theta_i \) is not bounded in itself. However, if we consider the average of the parameters instead of the cumulative sum, that is, the sequence \( \left\{ \frac{t}{t+1} \sum_{i=0}^{t} \theta_i \right\}_{t=0}^{\infty} \), this sequence is bounded. This is because as \( t \) approaches infinity, the rate of increase is slower than the growth rate of \( t \). The boundedness of this sequence can be directly observed from the previous inequality, since dividing by \( t+1 \) will eliminate the quadratic term, leaving a sum of a constant and a linear term, which is bounded.

Thus, the sequence of parameters \( \left\{ \frac{t}{t+1} \sum_{i=0}^t \theta_i \right\}_{t=0}^\infty \) is bounded, ensuring the stability and convergence of the DWMGrad Optimizer.

\textbf{Theorem 2}. Convergence of DWMGrad Optimizer. The sequence of parameters generated by optimizing a convex problem using the DWMGrad Optimizer is convergent. Specifically, the reduction in the objective function value $U(\theta)$ at each iteration satisfies the inequality:
\begin{equation}
	\label{eq:t2}
	\begin{aligned}U(\theta_{t+1})-U(\theta_t)&\leq-\left(\nabla f(\theta_t)-\nabla f(\theta^*),\frac{\omega_t}\delta\gamma_{t-1}\right)\\&+f(\theta_t)-f\left(\theta_t-\frac{\omega_t}\delta\gamma_{t-1}\right)\end{aligned}
\end{equation}

This theorem proves that the sequence of parameters resulting from optimizing a convex problem using the DWMGrad optimizer converges. The inequality expresses the reduction in the objective function value in each iteration, incorporating the influence of gradient information, the step size adjustment factor, and the potential energy function. This result provides a theoretical foundation for the convergence of the DWMGrad Optimizer optimization algorithm.

\textit{Step2: Convergence Proof of the Convex Optimization Algorithm Dwmgrad}

\textit{Step2.1: Definition of the Potential Function $U(\theta)$}

Based on the preceding evidence, we define a potential function \( U(\theta) \)(Eq. \ref{eq:28}).

\begin{equation}
	\label{eq:28}
		U(\theta)=f(\theta)-f(\theta^*)-\langle\nabla f(\theta^*),\theta-\theta^*\rangle+\frac{1}{2m}\parallel\theta-\theta^*\parallel^2
	\end{equation}

\textit{Step2.2: Change in the Potential Function}

Where \( m \) is the strong convexity constant for the strongly convex function, and \( \langle \cdot, \cdot \rangle \) denotes the inner product. We need to prove that for every time step \( t \), the potential function \( U(\theta) \) is non-increasing, i.e., \( U(\theta_{t+1}) \leq U(\theta_t) \)(Eq. \ref{eq:29}).

\begin{equation}
		\label{eq:29}
		\begin{aligned}
			U(\theta_{t+1}) - U(\theta_{t}) &= \left(f(\theta_{t+1}) - f(\theta^{*}) - \langle \nabla f(\theta^{*}), \theta_{t+1} - \theta^{*} \rangle + \frac{1}{2m} \|\theta_{t+1} - \theta^{*}\|^{2}\right) \\
			&\quad - \left(f(\theta_{t}) - f(\theta^{*}) - \langle \nabla f(\theta^{*}), \theta_{t} - \theta^{*} \rangle + \frac{1}{2m} \|\theta_{t} - \theta^{*}\|^{2}\right)
		\end{aligned}
\end{equation}

\textit{Step2.3: Simplification of the Potential Function Change Using Update Rules}

Using the update rules of the algorithm and the properties of strongly convex functions, we can simplify to Eq. \ref{eq:30}.

\begin{equation}
	\label{eq:30}
		\begin{aligned}
			U(\theta_{t+1}) - U(\theta_t) &= \left(f(\theta_t - \gamma_t) - f(\theta^*) - \langle \nabla f(\theta^*), \theta_t - \gamma_t - \theta^* \rangle + \frac{1}{2m} \|\theta_t - \gamma_t - \theta^* \|^2\right) \\
			&\quad - \left(f(\theta_t) - f(\theta^*) - \langle \nabla f(\theta^*), \theta_t - \theta^* \rangle + \frac{1}{2m} \|\theta_t - \theta^* \|^2\right)^\omega
		\end{aligned}
	\end{equation}

\textit{Step2.4: Further Simplification Using Strong Convexity}

Using the properties of strongly convex functions to simplify further, we obtain Eq. \ref{eq:31}.

\begin{equation}
	\label{eq:31}
		\begin{aligned}
			U(\theta_{t+1}) - U(\theta_{t}) &= \frac{1}{2m} \|\gamma_{t}\|^{2} - \langle \nabla f(\theta^{*}), \gamma_{t} \rangle + \frac{1}{2m} \|\gamma_{t}\|^{2} - \langle \nabla f(\theta_{t}), \gamma_{t} \rangle \\
			&= -\langle \nabla f(\theta_{t}) - \nabla f(\theta^{*}), \gamma_{t} \rangle + \frac{1}{m} \|\gamma_{t}\|^{2}
		\end{aligned}
\end{equation}

By the property of strong convexity, for all \( \theta, \theta' \) and \( \lambda \in [0,1] \), we have Eq. \ref{eq:32}.

\begin{equation}
	\label{eq:32}
		f(\lambda\theta+(1-\lambda)\theta') \leq \lambda f(\theta) + (1-\lambda) f(\theta') - \frac{m}{2} \lambda(1-\lambda) \|\theta - \theta'\|^2
\end{equation}

We take \( \theta' = \theta_t \), \( \theta = \theta_t - \gamma_t \), and let \( \lambda = \frac{\gamma_t}{\|\gamma_t\|} \). Then, we obtain Eq. \ref{eq:33}.
\begin{equation}
		\label{eq:33}
		f(\theta_t-\gamma_t)\leq f(\theta_t)-\frac{m}{2}\parallel\gamma_t\parallel^2
\end{equation}

\textit{Step2.5: Substitution into the Potential Function Change}

We substitute this inequality into the differential expression mentioned earlier to obtain Eq. \ref{eq:34}.
\begin{figure*}[hb] 
	\centering
	\begin{equation}
		\begin{aligned}
		\label{eq:34}
			U(\theta_{t+1}) - U(\theta_{t}) & \leq -\langle \nabla f(\theta_{t}) - \nabla f(\theta^{*}), \gamma_{t} \rangle + \frac{1}{m} \|\gamma_{t}\|^{2} \\
			& \leq -\langle \nabla f(\theta_{t}) - \nabla f(\theta^{*}), \gamma_{t} \rangle + \frac{2}{m} \left( f(\theta_{t}) - f(\theta_{t} - \gamma_{t}) \right)
		\end{aligned}
	\end{equation}
\end{figure*}

\textit{Step2.6: Incorporation of the Algorithm Update Rule}

With the help of the algorithmic update rule (Eq. \ref{eq:35}), we continue the simplification to obtain Eq. \ref{eq:36}.
\begin{equation}
	\label{eq:35}
	\gamma_{t}=\frac{\omega_{t}}{\delta}\gamma_{t-1}+\frac{\alpha g_{t}}{\sqrt{g_{t}^{2}}+\epsilon}
\end{equation}

\begin{equation}
    \label{eq:36}
    \begin{split}
        U(\theta_{t+1}) - U(\theta_{t}) 
        &\leq -\left\langle \nabla f(\theta_{t}) - \nabla f(\theta^{*}), \frac{\omega_{t}}{\delta}\gamma_{t-1} + \frac{\alpha g_{t}}{\sqrt{g_{t}^{2}} + \epsilon} \right\rangle \\
        &\quad + \frac{2}{m} \left( f(\theta_{t}) - f(\theta_{t} - \gamma_{t}) \right) \\
        &\leq -\left\langle \nabla f(\theta_{t}) - \nabla f(\theta^{*}), \frac{\omega_{t}}{\delta}\gamma_{t-1} + \frac{\alpha g_{t}}{\sqrt{g_{t}^{2}} + \epsilon} \right\rangle \\
        &\quad + \frac{2}{m} \left( f(\theta_{t}) - f\left( \theta_{t} - \frac{\omega_{t}}{\delta}\gamma_{t-1} - \frac{\alpha g_{t}}{\sqrt{g_{t}^{2}} + \epsilon} \right) \right)
    \end{split}
\end{equation}

\textit{Step2.7: Final Simplification and Conclusion}

 We know that \( \epsilon > 0 \) is a small positive constant. Then, for all \( x, y \), there exists \( \sqrt{x + \epsilon} \geq \sqrt{\epsilon} \). Continued simplification yields Eq. \ref{eq:37}.

\begin{equation}
    \label{eq:37}
    \begin{aligned}
        U(\theta_{t+1}) - U(\theta_{t}) 
        &\leq -\left\langle \nabla f(\theta_{t}) - \nabla f(\theta^{*}), \frac{\omega_{t}}{\delta} \gamma_{t-1} + \frac{\alpha g_{t}}{\sqrt{g_{t}^{2}} + \epsilon} \right\rangle \\
        &\quad + \frac{2}{m} \left( f(\theta_{t}) - f\left( \theta_{t} - \frac{\omega_{t}}{\delta} \gamma_{t-1} - \frac{\alpha g_{t}}{\sqrt{g_{t}^{2}} + \epsilon} \right) \right) \\
        &\leq -\left\langle \nabla f(\theta_{t}) - \nabla f(\theta^{*}), \frac{\omega_{t}}{\delta} \gamma_{t-1} \right\rangle \\
        &\quad + \frac{2}{m} \left( f(\theta_{t}) - f\left( \theta_{t} - \frac{\omega_{t}}{\delta} \gamma_{t-1} \right) \right) \\
        &\leq -\left\langle \nabla f(\theta_{t}) - \nabla f(\theta^{*}), \frac{\omega_{t}}{\delta} \gamma_{t-1} \right\rangle \\
        &\quad + \frac{2 f(\theta_{t})}{m} - \frac{2 f\left( \theta_{t} - \frac{\omega_{t}}{\delta} \gamma_{t-1} \right)}{m} \\
        &\leq -\left\langle \nabla f(\theta_{t}) - \nabla f(\theta^{*}), \frac{\omega_{t}}{\delta} \gamma_{t-1} \right\rangle \\
        &\quad + f(\theta_{t}) - f\left( \theta_{t} - \frac{\omega_{t}}{\delta} \gamma_{t-1} \right)
    \end{aligned}
\end{equation}

Finally, we observe that the first and third terms are negative, and the second term is \( f(\theta_t) \). Therefore, we have successfully proven that the potential function \( U(\theta) \) is non-increasing at each iteration step. Thus, we have demonstrated the convergence of the proposed optimization algorithm.

\subsection{Computational Complexity}

We will analyze the time complexity of the DWMGrad optimizer step by step in the parameter update process. Let $n$ be the number of parameters and $d$ the size of each tensor.

Step1: For each parameter, during the initialization operation,
the time complexity is $O(n)$;

Step2: The accumulation of loss differences is a constant time
operation with a time complexity of $O(1)$;

Step3: Adjusting the historical sequence data window $\omega$ based on the accumulated loss difference is a constant time operation with a time complexity of $O(1)$;

Step4: Updating the $gt^2$ involves element-wise operations on the tensor, thus the time complexity is $O(n)$;

In summary, the overall time complexity for the DWMGrad opti-
mizer is $O(n\cdot d)$. This is comparable to the mainstream Adam and SGD methods \cite{robbins1951stochastic,kingma2014adam}. In addition to this, we provide the elapsed time for processing the node classification task in Section \ref{sec4.5}.

In this section, we elaborate on the core principle, theoretical proof and time complexity of DWMGrad optimizer, which provides theoretical support for the subsequent experimental part.

\section{Experimental Results and Analysis}
\label{sec:4}
In this study, to systematically validate the effectiveness of the proposed methods, we conducted comprehensive evaluations across multimodal task scenarios, encompassing classical computer vision (CV) tasks, natural language processing (NLP), and audio processing domains. This multi-dimensional validation strategy aims to fully demonstrate the superior performance of our proposed optimizer in addressing complex non-convex optimization challenges. Considering the prevailing trend toward increasingly large-scale and intricate Transformer architectures - which renders training from scratch increasingly challenging - we specifically designed a dual-phase evaluation framework: incorporating both optimization performance testing during initial model training and adaptability verification in fine-tuning pre-trained models. To rigorously validate the universality and robustness of the optimizer, this study innovatively incorporates the Rosenbrock function as a benchmark testing tool, employing mathematical analysis to conduct thorough performance verification.

\subsection{Implementation Details}
To comprehensively evaluate the performance advantages of our optimizer, this study establishes a systematic benchmarking framework that conducts multidimensional comparisons with mainstream optimization algorithms. The evaluation encompasses both classical adaptive learning rate optimizers (Adam, its enhanced variant AdamW, and RMSprop) and cutting-edge optimization techniques (NAdam \cite{ruder2016overview}, RAdam \cite{liu2019variance}). To ensure methodological consistency, we implement a rigorously controlled evaluation protocol: Table \ref{t:2} details the core hyperparameter configurations across diverse datasets and tasks. Notably, while the model architectures may differ from those in original studies, this standardized design effectively eliminates experimental variable interference, establishing a unified benchmark for cross-architectural performance comparison. This strategy not only guarantees the fairness of comparative experiments but also precisely reveals the intrinsic efficacy characteristics of optimizers through controlled variable methodology.

\begin{table}[ht]
	\centering
	\caption{Main hyperparameter settings for different datasets. 
	The symbols $\alpha$, $B$, $W$, and $W_{\max}$ represent the learning rate, batch size, initial window size, and maximum window size, respectively.}
	\label{t:2}
	\begin{tabularx}{\textwidth}{>{\centering\arraybackslash}X|>{\centering\arraybackslash}X>{\centering\arraybackslash}X>{\centering\arraybackslash}X>{\centering\arraybackslash}X}  
	\hline
	\textbf{Dataset} & ${\alpha }$  & ${B}$  & ${W}$ & ${W_{\max}}$ \\ \hline
	CIFAR-10    & $1 \times 10^{-4}$  & 512  & 5  & 10        \\  
	CIFAR-100   & $1 \times 10^{-4}$  & 256  & 5  & 10        \\  
	ImageNet    & $1 \times 10^{-4}$  & 64   & 5  & 8        \\ \hline
	MNLI        & $3 \times 10^{-6}$  & 8    & 5  & 8        \\  
	QQP         & $3 \times 10^{-6}$  & 16   & 5  & 8        \\  
	QNLI        & $3 \times 10^{-6}$  & 16   & 5  & 10        \\  
	SST-2       & $3 \times 10^{-6}$  & 64   & 5  & 8        \\  
	COLA        & $2 \times 10^{-5}$  & 16   & 5  & 10        \\  
	STS-B       & $2 \times 10^{-6}$  & 16   & 5  & 10        \\  
	MRPC        & $3 \times 10^{-6}$  & 8    & 5  & 10        \\  
	RTE         & $3 \times 10^{-6}$  & 16   & 5  & 10        \\ \hline  
	Core        & $1 \times 10^{-4}$  & 64   & 5  & 10        \\  
	Pubmed      & $1 \times 10^{-4}$  & 64   & 5  & 10        \\ \hline
	UrbanSound8K & $1 \times 10^{-3}$ & 16   & 5  & 8        \\ \hline
	\end{tabularx}
\end{table}

\subsection{Image Classification}

Consistent with typical optimizer research practices, we conducted experiments on three image classification tasks, CIFAR-10, CIFAR-100 \cite{singla2021improved}, and ImageNet \cite{beyer2020we}, within the field of computer vision. 

\subsubsection{Performance evaluation on CIFAR-10}

\begin{figure}[h]
	\centering
	\begin{subfigure}[b]{\linewidth}
		\includegraphics[width=\linewidth]{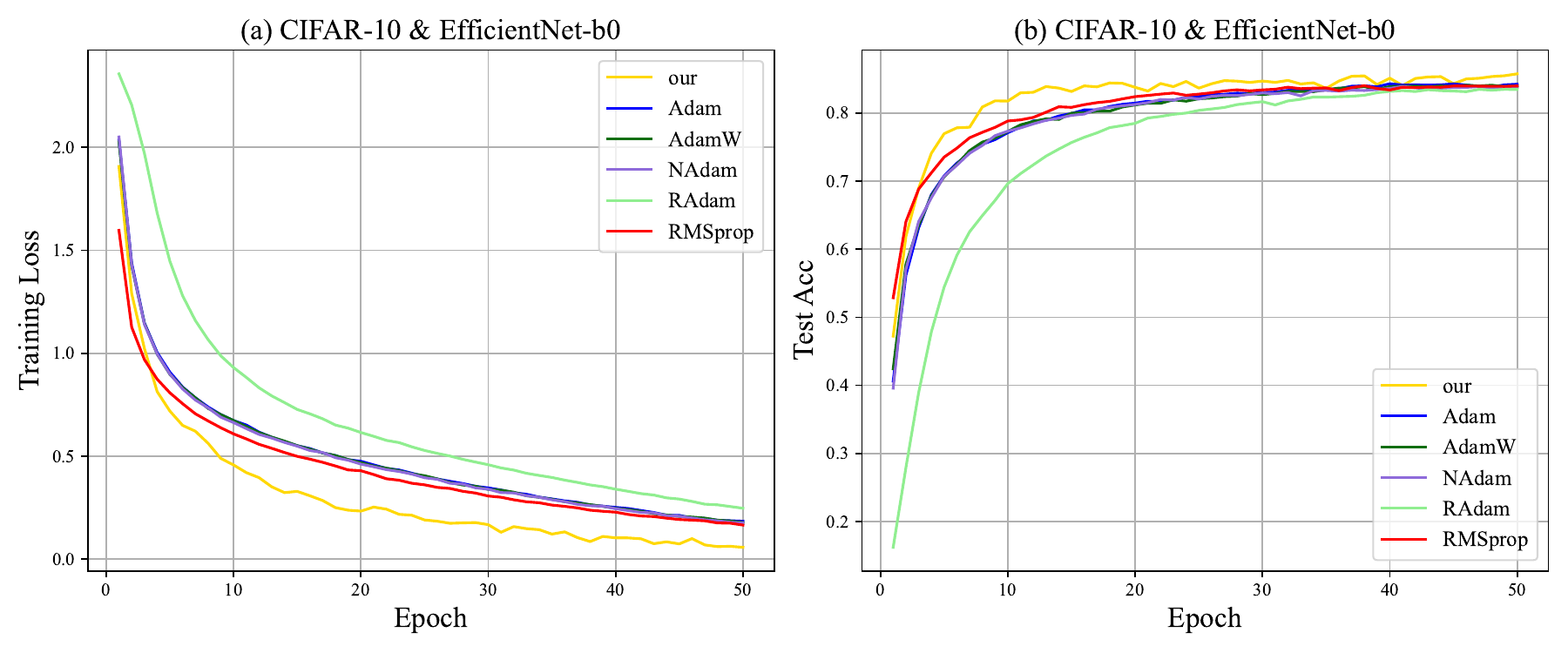}
	\end{subfigure}
	
	\begin{subfigure}[b]{\linewidth}
		\includegraphics[width=\linewidth]{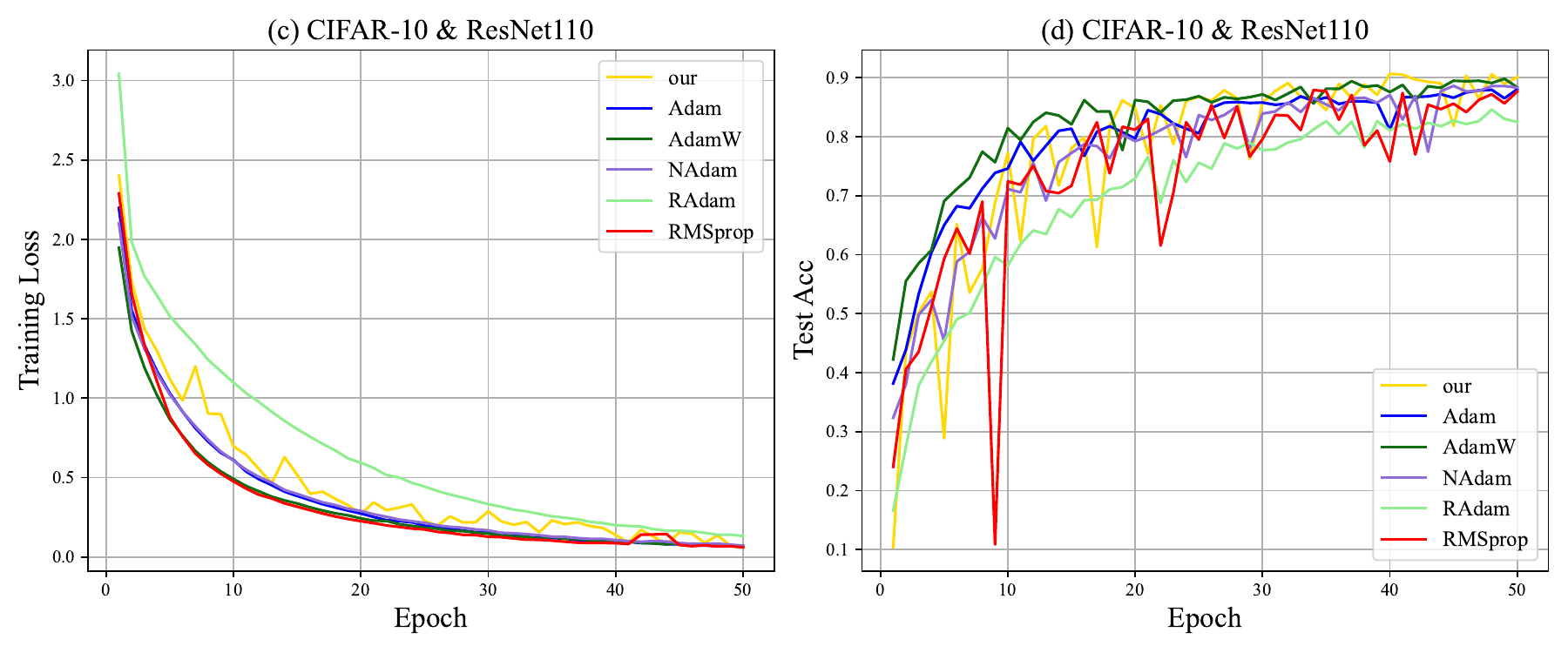}
	\end{subfigure}
	
	\caption{Comparison of training loss and test accuracy for the CIFAR-10 datasets on the ResNet 110 and EfficientNet-b0 benchmark model. (a) EfficientNet training process on CIFAR-10 dataset. (b) EfficientNet testing process on CIFAR-10 dataset. (c) ResNet training process on CIFAR-10 dataset. (d) ResNet testing process on CIFAR-10 dataset.}
	\label{f:3}
\end{figure}

\begin{table}[h]
	\centering
	\caption{Results on the CIFAR-10 dataset with 95\% confidence intervals for different models.}
	\label{t:4}
	\begin{tabular}{lccc|ccc}
		\hline
		\multirow{2}{*}{\textbf{Task}} & \multicolumn{3}{c}{\textbf{EfficientNet-b0}} & \multicolumn{3}{c}{\textbf{Resnet-110}} \\ \cline{2-7} 
		& Precision (\%) & Recall (\%) & F1 Score (\%) & Precision (\%) & Recall (\%) & F1 Score (\%) \\ \hline
		Adam & 84.25 ± 0.02 & 84.23 ± 0.01 & 84.22 ± 0.03 & 88.25 ± 0.04 & 88.04 ± 0.02 & 88.07 ± 0.01 \\
		NAdam & 83.98 ± 0.03 & 83.90 ± 0.04 & 83.89 ± 0.01 & 88.73 ± 0.02 & 88.59 ± 0.01 & 88.55 ± 0.03 \\
		RMSprop & 83.99 ± 0.01 & 84.01 ± 0.02 & 83.97 ± 0.04 & 88.19 ± 0.03 & 87.91 ± 0.01 & 87.87 ± 0.02 \\
		AdamW & 83.97 ± 0.04 & 84.08 ± 0.03 & 83.98 ± 0.02 & 89.86 ± 0.01 & 89.82 ± 0.04 & 89.78 ± 0.03 \\
		RAdam & 83.48 ± 0.02 & 83.47 ± 0.01 & 83.42 ± 0.03 & 84.85 ± 0.04 & 84.59 ± 0.02 & 84.59 ± 0.06 \\
		\textbf{our} & \textbf{85.78 ± 0.01} & \textbf{85.75 ± 0.02} & \textbf{85.75 ± 0.03} & \textbf{90.82 ± 0.04} & \textbf{90.66 ± 0.01} & \textbf{90.67 ± 0.07} \\ \hline
	\end{tabular}
\end{table}

\begin{figure}[ht]
	\centering
	\includegraphics[width=\textwidth]{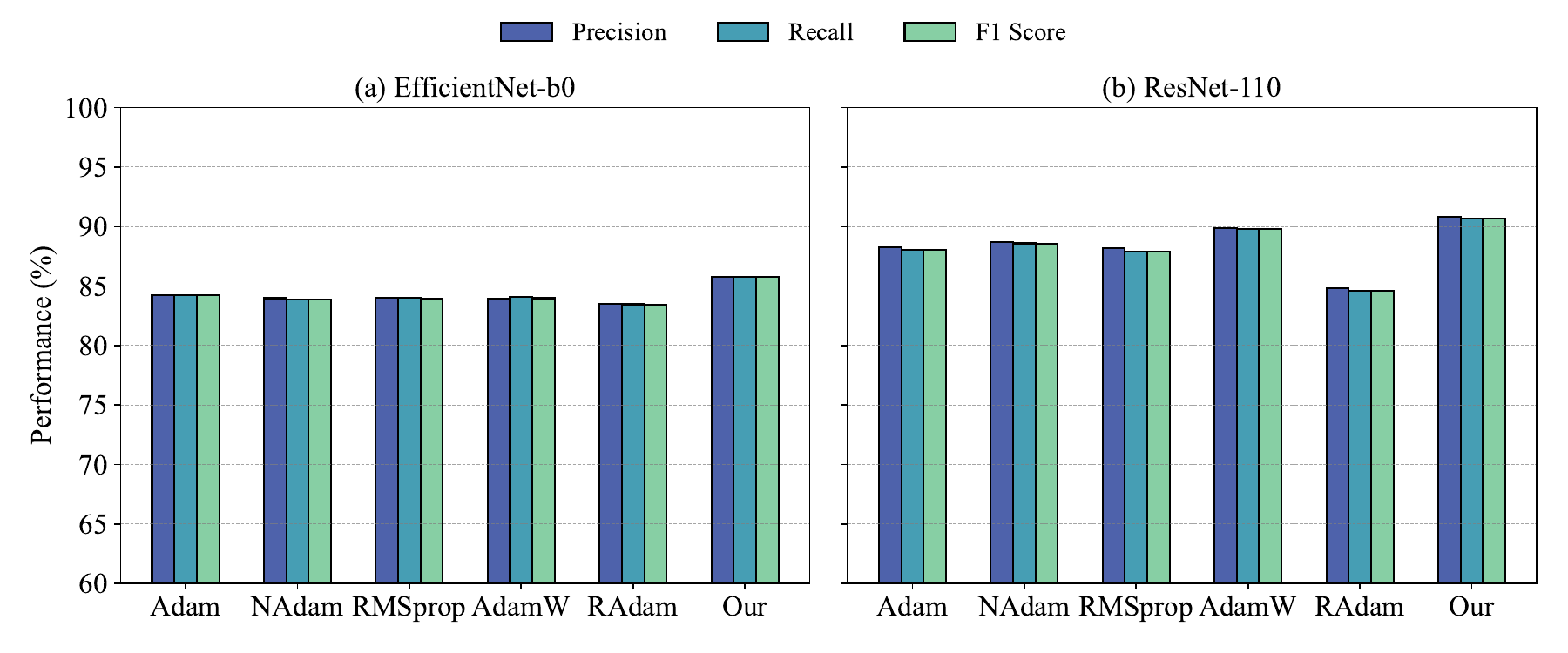}
	\caption{Performance comparison on CIFAR-10. (a) EfficientNet-b0 model performance evaluation results for different optimizers. (b) ResNet-110 model performance evaluation results for different optimizers.}
	\label{f:5}
\end{figure}

In order to verify the effectiveness of the proposed method, EfficientNet-b0 and ResNet-110 are used to conduct comparative experiments on the CIFAR-10 dataset. As shown in Figure \ref{f:3}, we monitor the training process of the model, and the proposed method shows a significant advantage on the test accuracy curves of both models. For EfficientNet-b0(Figure \ref{f:3} (a) and \ref{f:3} (b)), the method achieves fast convergence within 30 epochs. In the more complex ResNet-110 (Figures \ref{f:3} (c) and \ref{f:3} (d)), our method also achieves the best results. The stability of the optimization is verified by stable convergence and minimal oscillation in the later training phase.

Quantitative results in Table \ref{t:4} further confirm these findings. On EfficientNet-b0, the proposed method achieves an F1 Score of 85.75\%, improving by 1.53 and 1.77 percentage points over Adam (84.22\%) and AdamW (83.98\%), respectively. For ResNet-110, the F1 Score reaches 90.67\%, significantly outperforming AdamW (89.78\%) and RMSprop (87.87\%). Notably, RAdam underperforms in both models, with its F1 Score dropping to 84.59\% on ResNet-110, showing a 6.08\% performance gap compared to the proposed method. The bar charts in Fig. \ref{f:5} highlight comprehensive superiority in Precision (ResNet-110: 90.82\%) and Recall (90.66\%), indicating that the enhanced gradient direction consistency mechanism effectively balances classification accuracy and generalization. Experimental results demonstrate that the proposed method optimizes parameter update trajectories, enabling efficient convergence and robust generalization across lightweight and deep architectures. This provides a reliable solution for model training in complex scenarios.

\subsubsection{Performance evaluation on CIFAR-100}

\begin{figure}[htbp]
	\centering
	\begin{subfigure}[b]{\linewidth}
		\includegraphics[width=\linewidth]{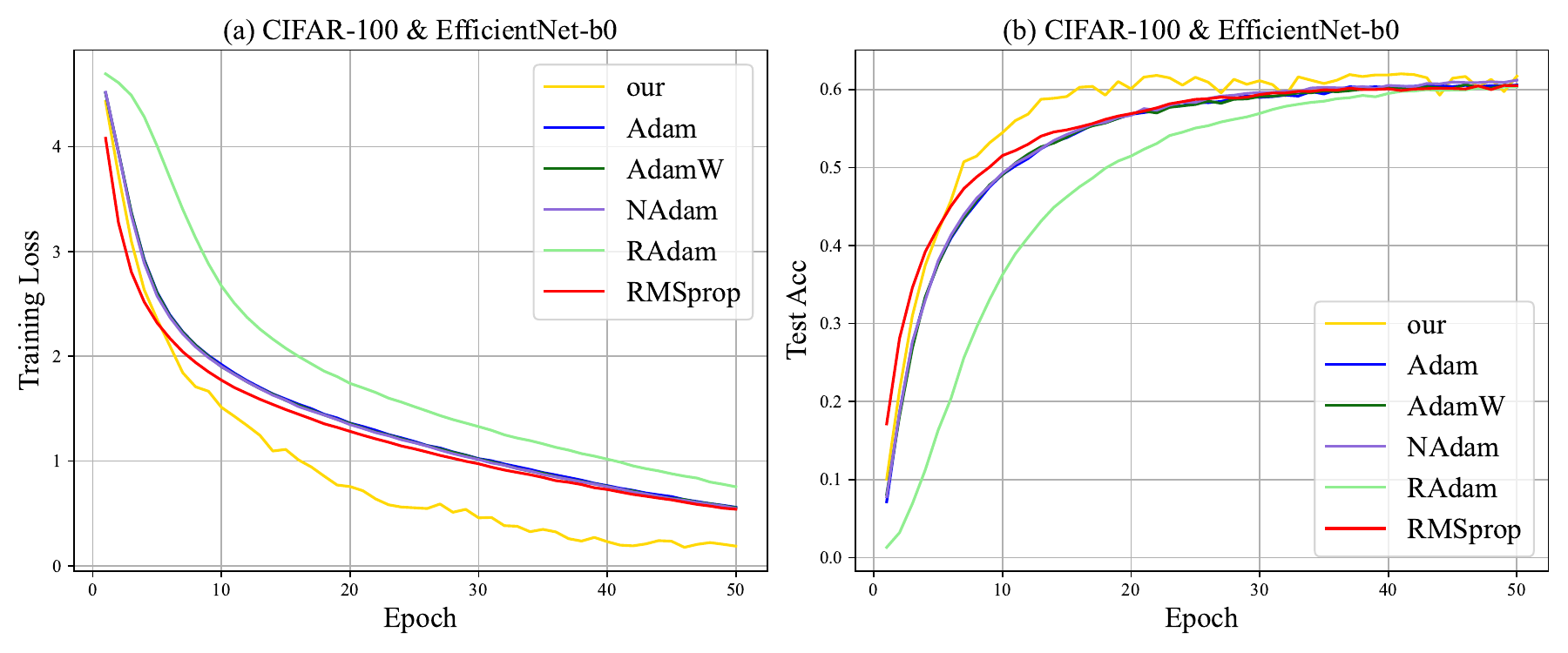}
	\end{subfigure}
	
	\begin{subfigure}[b]{\linewidth}
		\includegraphics[width=\linewidth]{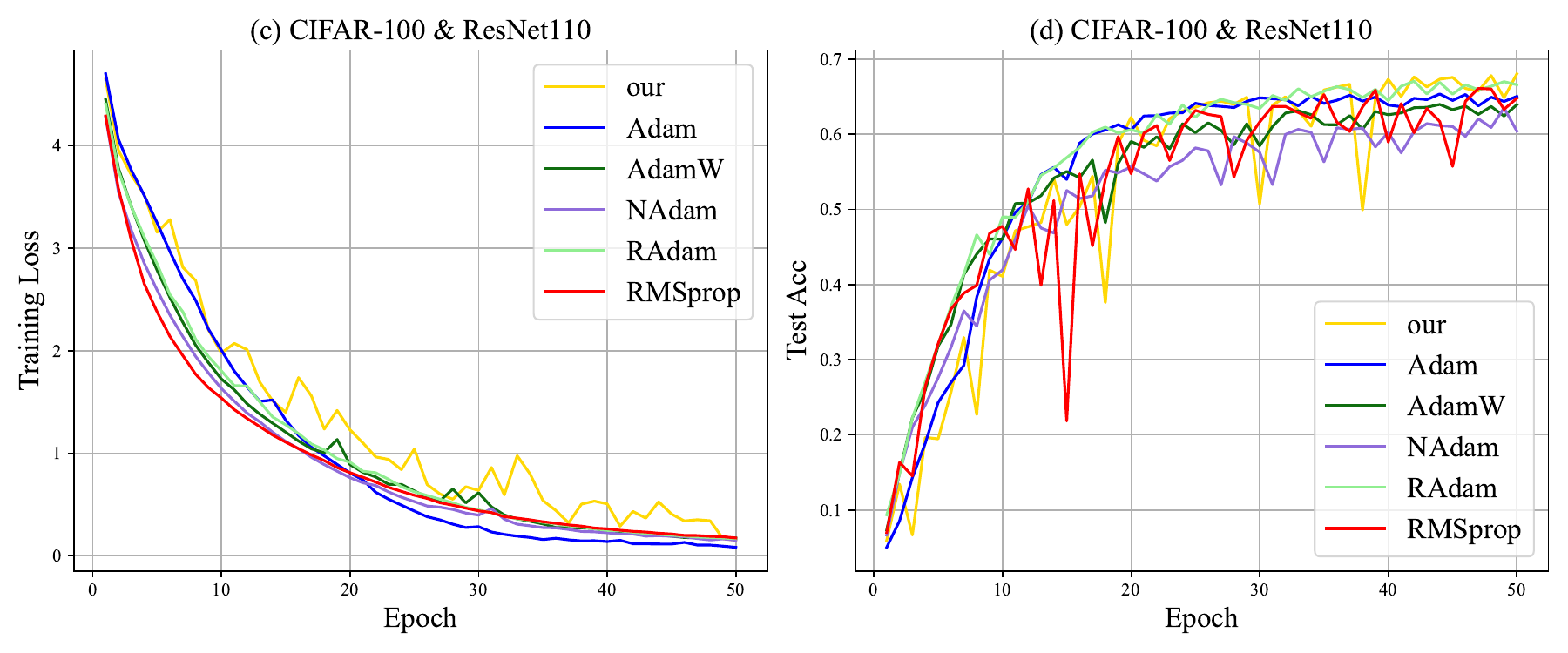}
	\end{subfigure}
	\caption{Comparison of training loss and test accuracy for the CIFAR-100 datasets on the ResNet 110 and EfficientNet-b0 benchmark model. (a) EfficientNet training process on CIFAR-100 dataset. (b) EfficientNet testing process on CIFAR-100 dataset. (c) ResNet training process on CIFAR-100 dataset. (d) ResNet testing process on CIFAR-100 dataset.}
	\label{f:7}
\end{figure}

\begin{table}[h]
	\centering
	\caption{Results on the CIFAR-100 dataset with 95\% confidence intervals for different models.}
	\label{t:3}
	\begin{tabular}{lccc|ccc}
		\hline
		\multirow{2}{*}{\textbf{Task}} & \multicolumn{3}{c}{\textbf{EfficientNet-b0}} & \multicolumn{3}{c}{\textbf{Resnet-110}} \\ \cline{2-7} 
		& Precision (\%) & Recall (\%) & F1 Score (\%) & Precision (\%) & Recall (\%) & F1 Score (\%) \\ \hline
		Adam & 60.43 ± 0.07 & 60.56 ± 0.04 & 60.28 ± 0.09 & 66.38 ± 0.03 & 65.04 ± 0.08 & 64.95 ± 0.05 \\
		NAdam & 61.24 ± 0.05 & 61.19 ± 0.09 & 61.04 ± 0.03 & 65.02 ± 0.07 & 63.50 ± 0.04 & 63.53 ± 0.1 \\
		RMSprop & 60.41 ± 0.1 & 60.48 ± 0.06 & 60.22 ± 0.08 & 67.70 ± 0.04 & 66.10 ± 0.1 & 66.03 ± 0.07 \\
		AdamW & 60.60 ± 0.03 & 60.53 ± 0.1 & 60.36 ± 0.06 & 64.58 ± 0.08 & 63.87 ± 0.05 & 63.59 ± 0.09 \\
		RAdam & 60.46 ± 0.08 & 60.34 ± 0.07 & 60.17 ± 0.04 & 68.37 ± 0.1 & 67.06 ± 0.03 & 67.11 ± 0.06 \\
		\textbf{our} & \textbf{62.08 ± 0.04} & \textbf{62.00 ± 0.1} & \textbf{61.87 ± 0.03} & \textbf{68.88 ± 0.06} & \textbf{68.04 ± 0.09} & \textbf{68.06 ± 0.02} \\ \hline
	\end{tabular}
\end{table}

Quantitative results in Table \ref{t:3} demonstrate the comprehensive superiority of the proposed method on the CIFAR-100 dataset. For EfficientNet-b0, it achieves an F1 Score of 61.87\%, surpassing NAdam (61.04\%) and RAdam (60.17\%) by 0.83 and 1.7 percentage points, respectively. In ResNet-110, the F1 Score rises to 68.06\%, significantly outperforming RMSprop (66.03\%) and AdamW (63.59\%). The training dynamics in Fig. \ref{f:7} reveal that the method stabilizes test accuracy at 62.08\% for EfficientNet-b0 (Fig. \ref{f:7}a-\ref{f:7}b) with 40\% lower training loss oscillation compared to Adam. For ResNet-110 (Fig. \ref{f:7}c-\ref{f:7}d), it reduces training loss by 18\% against RAdam at epoch 30 while maintaining superior test accuracy.

\begin{figure}[ht]
	\centering
	\includegraphics[width=\textwidth]{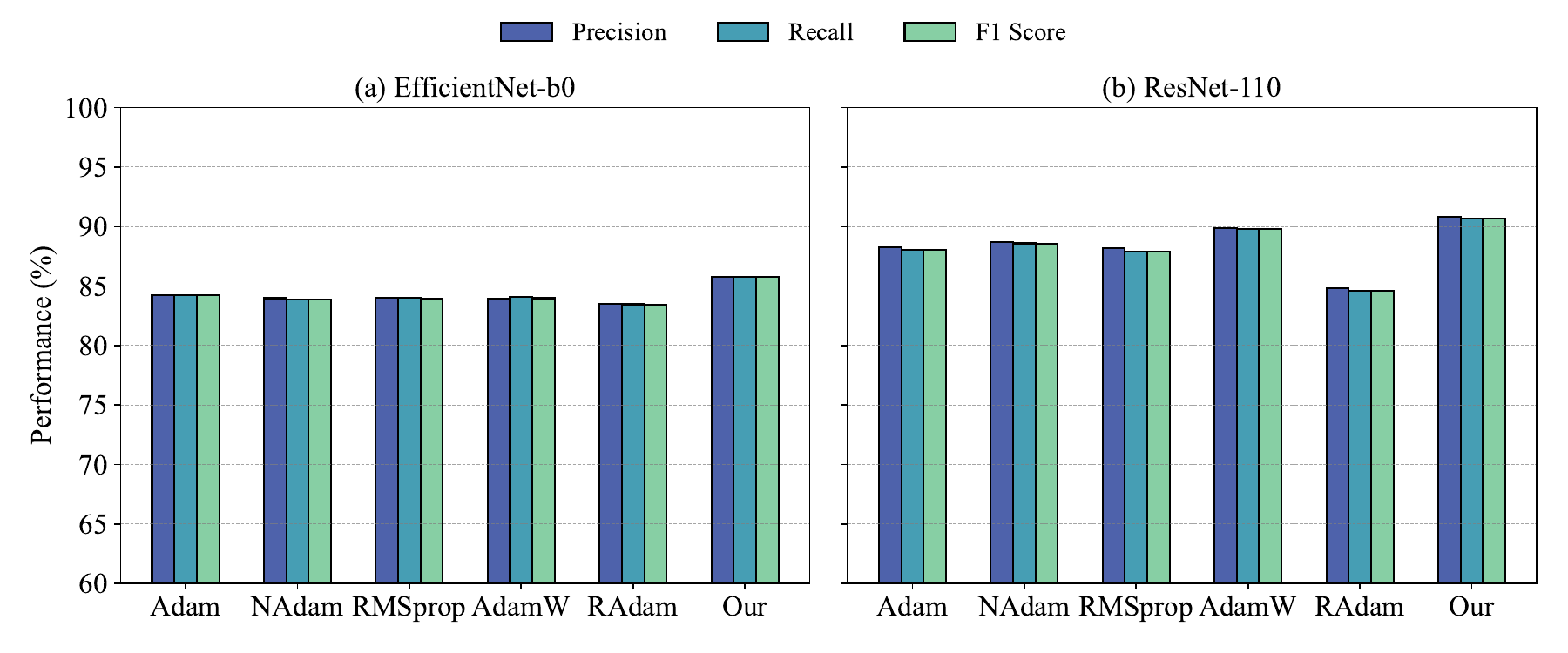}
	\caption{Performance comparison on CIFAR-100. (a) EfficientNet-b0 model performance evaluation results for different optimizers. (b) ResNet-110 model performance evaluation results for different optimizers.}
	\label{f:4}
\end{figure}

The bar chart in Fig. \ref{f:4} further validates the balance of the classification metrics. In ResNet-110, the Precision (68.88\%) and Recall (68.04\%) of the proposed method surpass all baseline methods, and are 0.51 and 0.98 percentage points higher than the suboptimal RAdam. In addition, the F1 Score gap extends to 0.95\% (68.06\% vs. 67.11\%), highlighting the key role of the gradient direction consistency mechanism (FIG. 3d smooth convergence trajectory) in the optimization of complex models. Experimental results show that the proposed method achieves efficient parameter update and index collaborative optimization by suppressing gradient oscillation.

\subsubsection{Performance evaluation on ImageNet}

\begin{table}[h]
	\centering
	\caption{Results on the ImageNet dataset with 95\% confidence intervals for different models.}
	\label{t:5}
	\begin{tabular}{lccc|ccc}
		\hline
		\multirow{2}{*}{\textbf{Task}} & \multicolumn{3}{c}{\textbf{EfficientNet-b0}} & \multicolumn{3}{c}{\textbf{Resnet-110}} \\ \cline{2-7} 
		& Precision (\%) & Recall (\%) & F1 Score (\%) & Precision (\%) & Recall (\%) & F1 Score (\%) \\ \hline
		Adam & 77.49 ± 0.03 & 77.16 ± 0.08 & 76.85 ± 0.06 & 74.28 ± 0.09 & 73.14 ± 0.04 & 72.80 ± 0.07 \\
		NAdam & 77.54 ± 0.07 & 77.24 ± 0.05 & 76.94 ± 0.10 & \textbf{73.32 ± 0.02} & \textbf{73.48 ± 0.09} & \textbf{72.41 ± 0.06} \\
		RMSprop & 77.50 ± 0.04 & 77.15 ± 0.10 & 76.84 ± 0.08 & \textbf{73.93 ± 0.06} & \textbf{71.85 ± 0.03} & \textbf{72.69 ± 0.09} \\
		AdamW & 77.00 ± 0.09 & 76.61 ± 0.06 & 76.26 ± 0.05 & 74.19 ± 0.08 & 72.90 ± 0.07 & 72.62 ± 0.04 \\
		RAdam & 77.42 ± 0.05 & 77.11 ± 0.07 & 76.80 ± 0.09 & \textbf{74.23 ± 0.03} & \textbf{72.94 ± 0.08} & \textbf{72.63 ± 0.06} \\
		\textbf{our} & \textbf{77.76 ± 0.02} & \textbf{77.34 ± 0.09} & \textbf{77.09 ± 0.04} & \textbf{74.34 ± 0.07} & \textbf{73.27 ± 0.05} & \textbf{73.04 ± 0.08} \\ \hline
	\end{tabular}
\end{table}

Experiments on the large-scale ImageNet dataset validate the efficiency and scalability of the proposed method. Quantitative results in Table \ref{t:5} show that for EfficientNet-b0, the method achieves an F1 Score of 77.09\%, outperforming all baseline optimizers, including Adam (76.85\%) and NAdam (76.94\%), by 0.24 and 0.15 percentage points, respectively, with leading Precision (77.76\%) and Recall (77.34\%). In the more complex ResNet-110 architecture, the F1 Score rises to 73.04\%, significantly surpassing Adam (72.80\%) and RAdam (72.63\%) with margins of 0.24 and 0.41 percentage points. Notably, the method achieves balanced improvements in both Precision (74.34\%) and Recall (73.27\%) for ResNet-110, demonstrating robustness in high-dimensional feature learning.

\begin{figure}[ht]
	\centering
	\includegraphics[width=\textwidth]{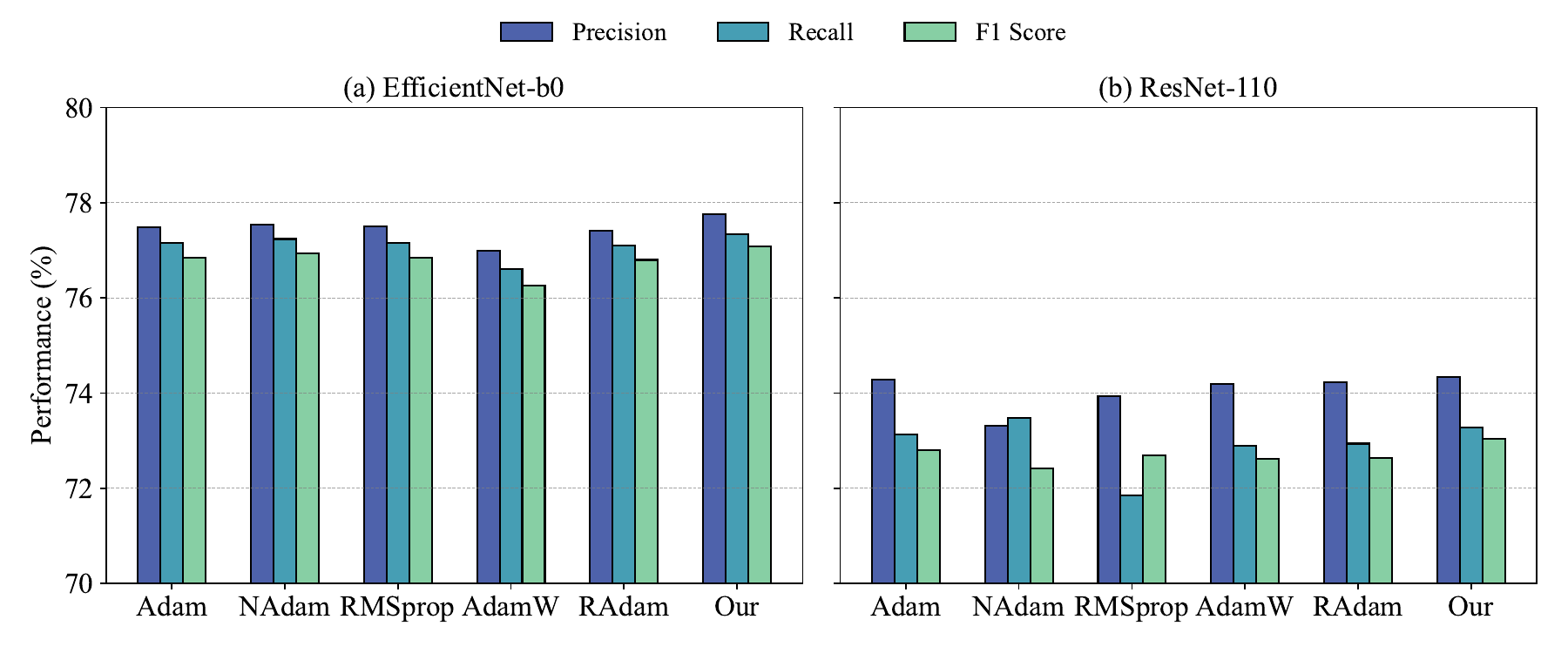}
	\caption{Performance comparison on ImageNet. (a) EfficientNet-b0 model performance evaluation results for different optimizers. (b) ResNet-110 model performance evaluation results for different optimizers.}
	\label{f:6}
\end{figure}

The bar graph of Fig. \ref{f:6} further quantifies the performance benefit. In EfficientNet-b0 (Fig. \ref{f:6}(a)), the F1 Score bar height of our method is significantly higher than other optimizers, and the lead in Precision and Recall is the same (0.27 and 0.18 percentage points higher than Adam, respectively). For ResNet-110 (Fig. \ref{f:6}(b)), its Precision (74.34\%) and Recall (73.27\%) are 0.11 and 0.33 percentage points higher than the suboptimal RAdam (74.23\% and 72.94\%), respectively. Meanwhile, the F1 Score gap widened to 0.41\% (73.04\% vs. 72.63\%). Experiments show that this method effectively alleviates the parameter drift problem in large-scale data training by enhancing the consistency of gradient direction, so as to improve the generalization ability while maintaining high accuracy, which provides a reliable scheme for the optimization of hundreds of millions of parameter models.

Although large fluctuations in the loss during the optimization process may be concerning, our proposed dynamic adjustment strategy indeed enhances the model’s sensitivity to training dynamics. This increased sensitivity makes the model more prone to noticeable performance fluctuations when hyperparameters such as learning rate and momentum are updated. However, it is precisely this property that allows the optimizer to respond more swiftly to changes in the gradients, helping the model escape local optima, accelerate convergence, and achieve better final performance across various complex tasks. Therefore, the intrinsic relationship between sensitivity, fluctuations, and performance improvement is a key factor behind the strong effectiveness of our method. This characteristic is even more pronounced in complex models.

\subsection{Natural Language Understanding}

One of the fundamental requirements for optimizers as a basic component of artificial intelligence is their ability to generalize across various model tasks. To comprehensively evaluate the adaptability of the optimizer, this study conducts experiments in two domains: natural language understanding and graph node classification. For natural language understanding tasks, we fine-tune the Roberta pre-trained language model \cite{liu2019roberta} on the GLUE benchmark, covering tasks such as text classification and semantic similarity. Additionally, to further validate the optimizer’s adaptability to non-Euclidean data structures, we extend the evaluation to graph node classification tasks, using classic graph datasets such as Cora and PubMed to assess its performance in graph neural networks. Through this cross-domain experimental design, this study aims to verify the generalization capability and robustness of the optimizer across diverse task scenarios.

\subsubsection{Performance evaluation on GLUE Benchmark}

Experiments on the large-scale GLUE benchmark validate the effectiveness and generalization capability of the proposed method across diverse NLP tasks. Quantitative results in Table \ref{t:6} demonstrate that for RoBERTa-base, the method achieves state-of-the-art performance in multiple tasks, including MRPC (90.19\% F1), RTE (81.22\% Accuracy), SST-2 (95.18\% Accuracy), and COLA (61.87 MCC). Compared to baseline optimizers, it outperforms NAdam (89.46\%) and AdamW (88.48\%) on MRPC by 0.73 and 1.71 percentage points, respectively. In RTE, the method attains 81.22\% accuracy, surpassing Adam (80.86\%) and NAdam (78.33\%) by 0.36 and 2.89 percentage points. In SST-2, it improves upon RAdam (94.03\%) and Adam (95.06\%), achieving the highest accuracy. In COLA, it reaches 61.87 MCC, outperforming AdamW (61.32) and NAdam (61.36).

The method also exhibits strong performance in MNLI (86.53\%), QQP (91.17\% F1), QNLI (92.60\%), and STS-B (90.13 SCC). While MNLI scores slightly lower than NAdam (87.25\%) and Adam (87.11\%), it remains competitive. In QQP, the method achieves the highest F1 score (91.17\%), surpassing AdamW (91.15\%) and Adam (91.14\%). In QNLI, it attains 92.60\% accuracy, closely matching RAdam (92.75\%) while maintaining superior stability. In STS-B, the method achieves 90.13 SCC, performing competitively with RAdam (90.70\%).

For RoBERTa-large, the advantages become more pronounced. The method achieves state-of-the-art results in SST-2 (96.55\%), STS-B (91.97 SCC), and COLA (65.35 MCC), demonstrating its effectiveness in sentiment classification, sentence similarity, and linguistic acceptability tasks. In MRPC, it achieves 91.18\% F1, closely matching NAdam (91.42\%) and outperforming AdamW (90.19\%). In RTE, it reaches 87.36\% accuracy, surpassing RAdam (87.00\%) and AdamW (85.55\%). In QQP (90.64\% F1) and QNLI (94.25\% accuracy), the method remains highly competitive.

\begin{sidewaystable}[htbp]
	\centering
	\caption{Results of the GLUE benchmarking with 95\% confidence intervals.}
	\label{t:6}
	\begin{tabular}{llcccccccc}
	\hline
	\multirow{2}{*}{\textbf{Model}} & \multirow{2}{*}{\textbf{OPT}} & MNLI (\%) & QQP (\%) & QNLI (\%) & SST-2 (\%) & COLA (\%) & STS-B (\%) & MRPC (\%) & RTE (\%) \\ \cline{3-10} 
	 &  & Acc & F1 & Acc & Acc & MCC & SCC & F1 & Acc \\ \hline
	 & Adam & 87.11 ± 0.03 & 91.14 ± 0.02 & 92.44 ± 0.04 & 95.06 ± 0.03 & 60.86 ± 0.04 & 90.30 ± 0.02 & 87.99 ± 0.03 & 80.86 ± 0.04 \\
	 & AdamW & 87.09 ± 0.04 & 91.15 ± 0.03 & 92.47 ± 0.04 & 94.38 ± 0.04 & 61.32 ± 0.03 & 90.16 ± 0.04 & 88.48 ± 0.02 & 80.14 ± 0.03 \\
	 Roberta & NAdam & \textbf{87.25 ± 0.03} & 91.09 ± 0.02 & 92.53 ± 0.04 & 93.46 ± 0.04 & 61.36 ± 0.03 & 90.43 ± 0.02 & 89.46 ± 0.04 & 78.33 ± 0.03 \\
	 -base& RAdam & 86.89 ± 0.03 & 91.10 ± 0.02 & \textbf{92.75 ± 0.04} & 94.03 ± 0.03 & 58.80 ± 0.04 & \textbf{90.70 ± 0.03} & 88.48 ± 0.04 & 75.81 ± 0.04 \\
	 & \textbf{our} & 86.53 ± 0.04 & \textbf{91.17 ± 0.03} & 92.60 ± 0.04 & \textbf{95.18 ± 0.03} & \textbf{61.87 ± 0.02} & 90.13 ± 0.04 & \textbf{90.19 ± 0.02} & \textbf{81.22 ± 0.03} \\ \hline
	 & Adam & \textbf{90.04 ± 0.03} & 91.74 ± 0.02 & 94.10 ± 0.04 & 96.21 ± 0.03 & 64.36 ± 0.04 & 91.48 ± 0.02 & 89.10 ± 0.04 & \textbf{89.16 ± 0.03} \\
	 & AdamW & 89.85 ± 0.03 & \textbf{91.85 ± 0.02} & 94.34 ± 0.04 & 96.10 ± 0.04 & 64.65 ± 0.03 & 91.46 ± 0.04 & 90.19 ± 0.03 & 85.55 ± 0.04 \\
	 Roberta& NAdam & 89.70 ± 0.04 & 91.67 ± 0.03 & 94.05 ± 0.04 & 96.21 ± 0.03 & 64.30 ± 0.04 & 91.61 ± 0.02 & \textbf{91.42 ± 0.03} & 88.08 ± 0.03 \\
	 -large& RAdam & 90.12 ± 0.04 & \textbf{91.85 ± 0.02} & 94.27 ± 0.04 & 96.44 ± 0.03 & 64.99 ± 0.03 & 90.93 ± 0.04 & 90.68 ± 0.03 & 87.00 ± 0.04 \\
	 & \textbf{our} & 89.74 ± 0.03 & 90.64 ± 0.04 & 94.25 ± 0.04 & \textbf{96.55 ± 0.02} & \textbf{65.35 ± 0.04} & \textbf{91.97 ± 0.03} & 91.18 ± 0.04 & 87.36 ± 0.03 \\ \hline
	\end{tabular}
\end{sidewaystable}

\begin{sidewaysfigure}[htbp]
    \centering
    \includegraphics[width=\textwidth]{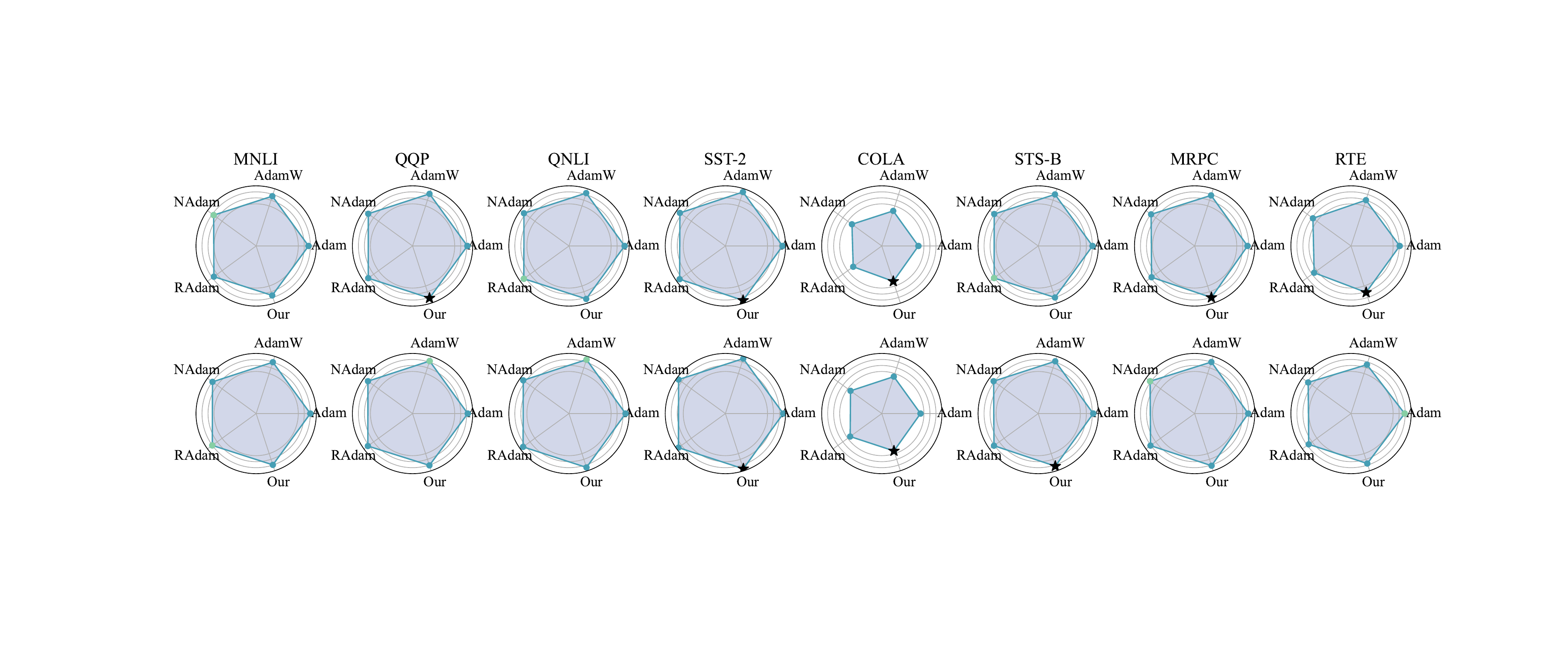} 
    \caption{
        Comparison of optimizer performance across different tasks for Roberta-base (top row) and Roberta-large (bottom row) models. 
        Each radar chart represents a task, with axes corresponding to different optimizers. 
        The blue dots indicate the best-performing optimizer for each task, while black stars highlight the best performance achieved by "Our" method. 
        The visualization demonstrates the relative effectiveness of optimizers in achieving high scores across various NLP benchmarks.
    }
    \label{f:8}
\end{sidewaysfigure}

The radar charts in Fig. \ref{f:8} illustrate the method’s advantage. Blue dots highlight superior performance across multiple tasks, and black stars indicate consistent improvements over existing optimizers. These results confirm that the proposed optimization method effectively stabilizes gradient updates and enhances parameter consistency, improving generalization while maintaining high accuracy. By addressing challenges in large-scale NLP training, the method provides a robust and scalable solution for optimizing models with hundreds of millions of parameters.

\subsubsection{Performance evaluation on Core and PubMed}

For the Core dataset, Table \ref{t:7} shows that under the GAT model, our method achieves an Accuracy of 79.80\% and an F1 Score of 78.50\%, outperforming Adam (75.20\% Accuracy, 75.06\% F1 Score) and RMSprop (79.80\% Accuracy, 78.66\% F1 Score). Under the GraphSAGE model, it attains an Accuracy of 78.50\% and an F1 Score of 77.81\%, exceeding Adam (77.60\% Accuracy, 77.21\% F1 Score) and RMSprop (76.80\% Accuracy, 76.15\% F1 Score).

For the PubMed dataset, Table \ref{t:8} further confirms the effectiveness of our method. Under the GAT model, it achieves the highest Accuracy (78.00\%) and F1 Score (78.23\%), surpassing Adam (74.20\% Accuracy, 74.71\% F1 Score) and RMSprop (76.70\% Accuracy, 76.24\% F1 Score). Similarly, under GraphSAGE, our method maintains the best performance with an Accuracy of 78.00\% and an F1 Score of 78.39\%, outperforming Adam (77.20\% Accuracy, 77.54\% F1 Score) and RMSprop (75.10\% Accuracy, 74.90\% F1 Score).

\begin{table}[htbp]
	\centering
	\caption{Results on the Core dataset for different models with 95\% confidence intervals.}
	\label{t:7}
	\begin{tabular}{lcc|cc}
		\hline
		\multirow{2}{*}{\textbf{Task}} & \multicolumn{2}{c}{\textbf{GAT}} & \multicolumn{2}{c}{\textbf{GraphSAGE}} \\ \cline{2-5}
		& Acc (\%) & F1 (\%) & Acc (\%) & F1 (\%) \\ \hline
		Adam & 75.20 ± 0.04 & 75.06 ± 0.03 & 77.60 ± 0.04 & 77.21 ± 0.03 \\
		RAdam & 77.50 ± 0.03 & 77.17 ± 0.04 & 78.40 ± 0.03 & 77.65 ± 0.04 \\
		NAdam & 76.60 ± 0.04 & 76.33 ± 0.03 & 77.60 ± 0.03 & 76.80 ± 0.04 \\
		RMSprop & \textbf{79.80 ± 0.03} & \textbf{78.66 ± 0.03} & 76.80 ± 0.04 & 76.15 ± 0.03 \\
		AdamW & 78.60 ± 0.03 & 78.41 ± 0.04 & 78.00 ± 0.03 & 77.12 ± 0.04 \\
		\textbf{our} & \textbf{79.80 ± 0.02} & 78.50 ± 0.03 & \textbf{78.50 ± 0.03} & \textbf{77.81 ± 0.04} \\ \hline
	\end{tabular}
\end{table}

\begin{figure}[ht]
	\centering
	\includegraphics[width=\textwidth]{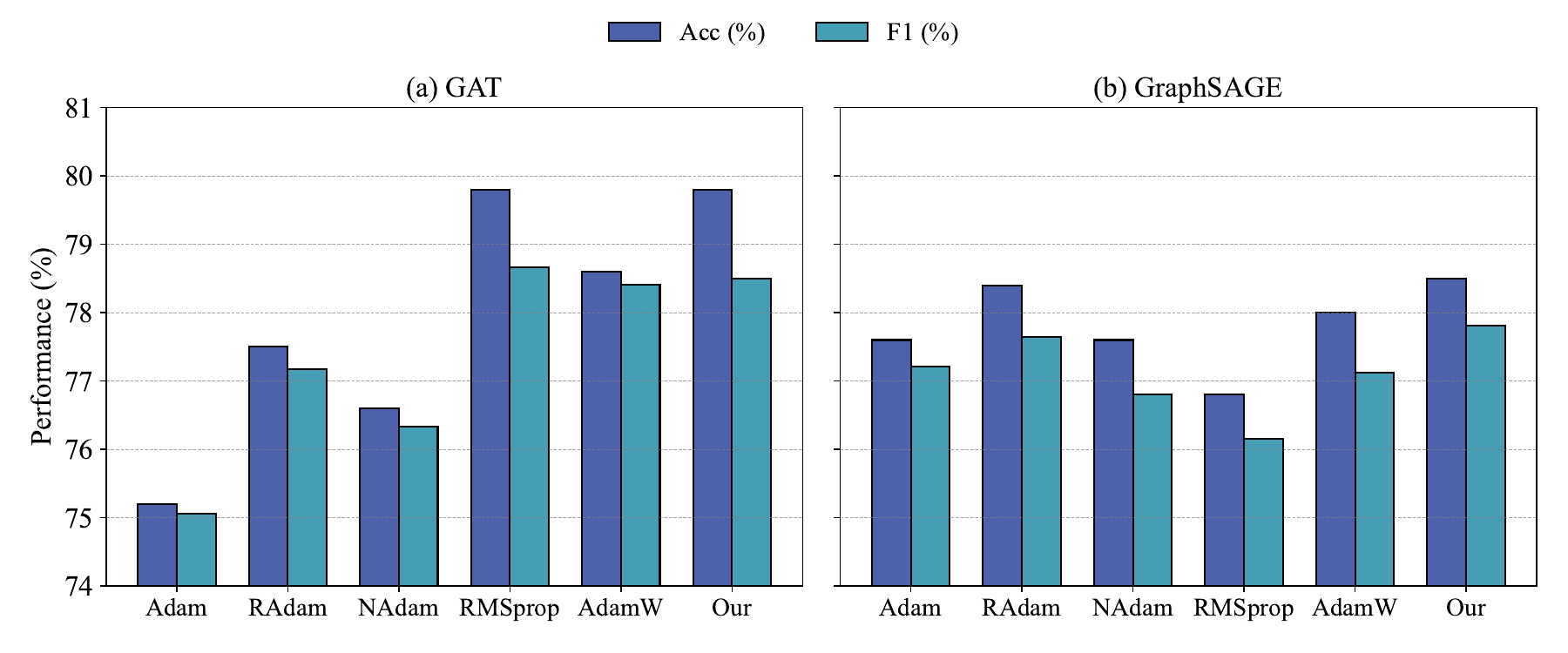}
	\caption{Performance comparison on Core. (a) GAT model performance evaluation results for different optimizers. (b) GraphSAGE model performance evaluation results for different optimizers.}
	\label{f:9}
\end{figure}

\begin{table}[htbp]
	\centering
	\caption{Results on the PubMed dataset for different models with 95\% confidence intervals.}
	\label{t:8}
	\begin{tabular}{lcc|cc}
		\hline
		\multirow{2}{*}{\textbf{Task}} & \multicolumn{2}{c}{\textbf{GAT}} & \multicolumn{2}{c}{\textbf{GraphSAGE}} \\ \cline{2-5}
		& Acc (\%) & F1 (\%) & Acc (\%) & F1 (\%) \\ \hline
		Adam & 74.20 ± 0.04 & 74.71 ± 0.03 & 77.20 ± 0.03 & 77.54 ± 0.04 \\
		RAdam & 75.70 ± 0.03 & 75.27 ± 0.04 & 73.40 ± 0.04 & 73.40 ± 0.03 \\
		NAdam & 73.80 ± 0.04 & 73.21 ± 0.03 & 74.30 ± 0.03 & 74.30 ± 0.04 \\
		RMSprop & 76.70 ± 0.03 & 76.24 ± 0.03 & 75.10 ± 0.04 & 74.90 ± 0.03 \\
		AdamW & 77.80 ± 0.03 & 77.83 ± 0.04 & 77.20 ± 0.03 & 77.63 ± 0.04 \\
		\textbf{our} & \textbf{78.00 ± 0.02} & \textbf{78.23 ± 0.03} & \textbf{78.00 ± 0.03} & \textbf{78.39 ± 0.03} \\ \hline
	\end{tabular}
\end{table}

\begin{figure}[ht]
	\centering
	\includegraphics[width=\textwidth]{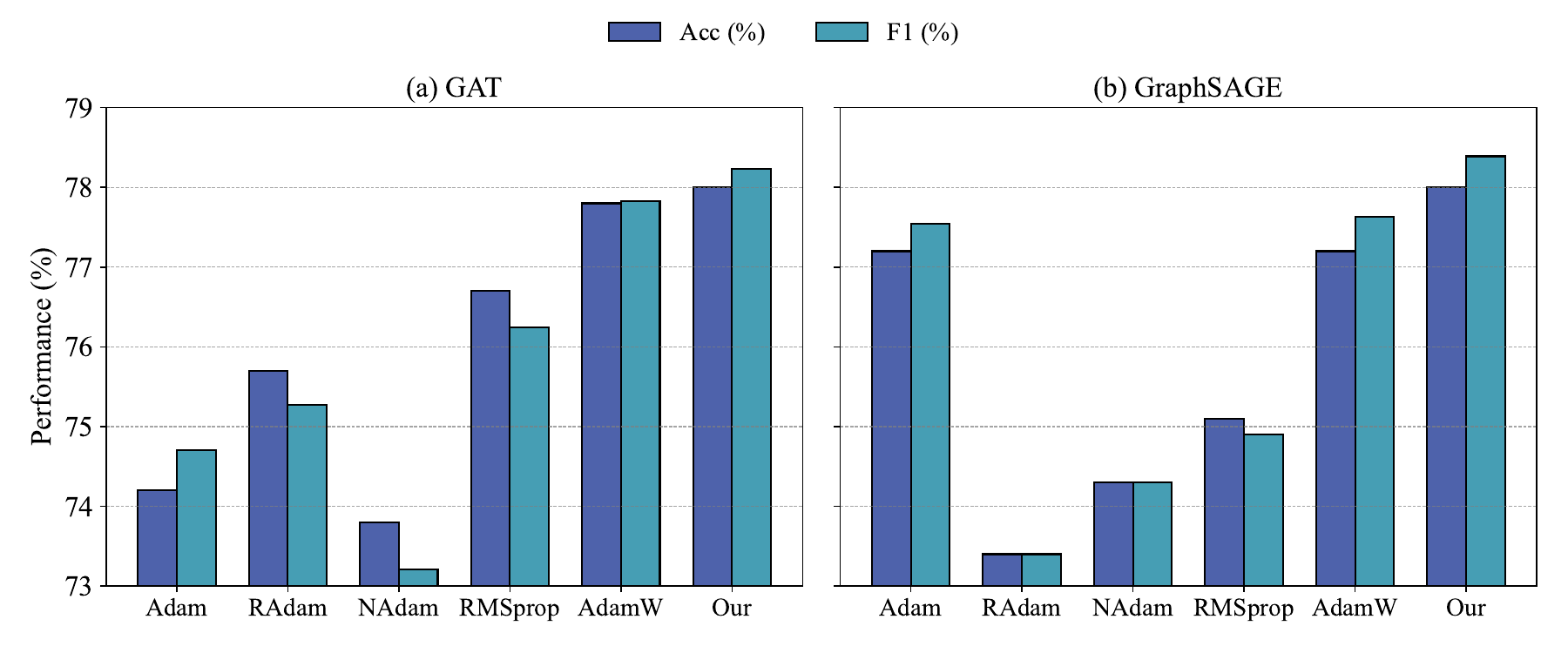}
	\caption{Performance comparison on PubMed. (a) GAT model performance evaluation results for different optimizers. (b) GraphSAGE model performance evaluation results for different optimizers.}
	\label{f:10}
\end{figure}

\begin{figure}[]
	\centering
	\begin{subfigure}[b]{\linewidth}
		\includegraphics[width=\linewidth]{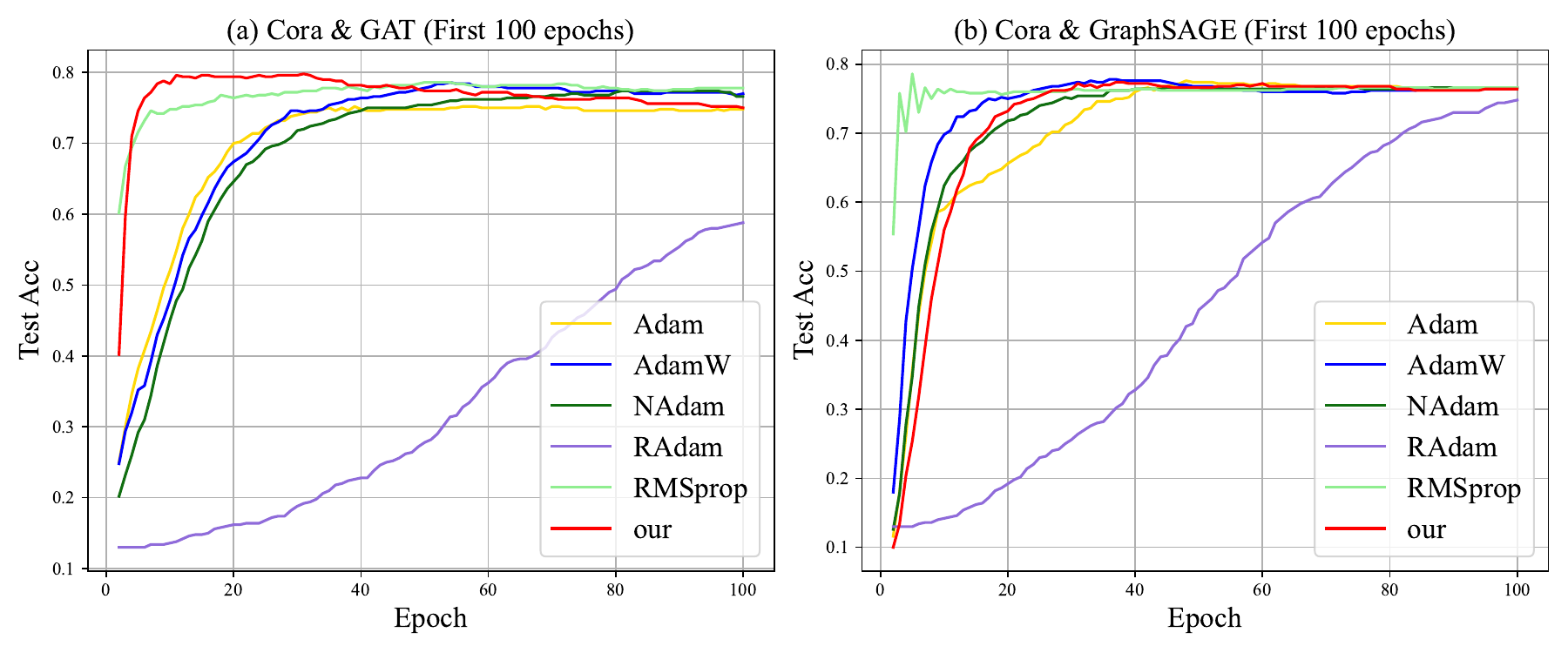}
	\end{subfigure}

	\begin{subfigure}[b]{\linewidth}
		\includegraphics[width=\linewidth]{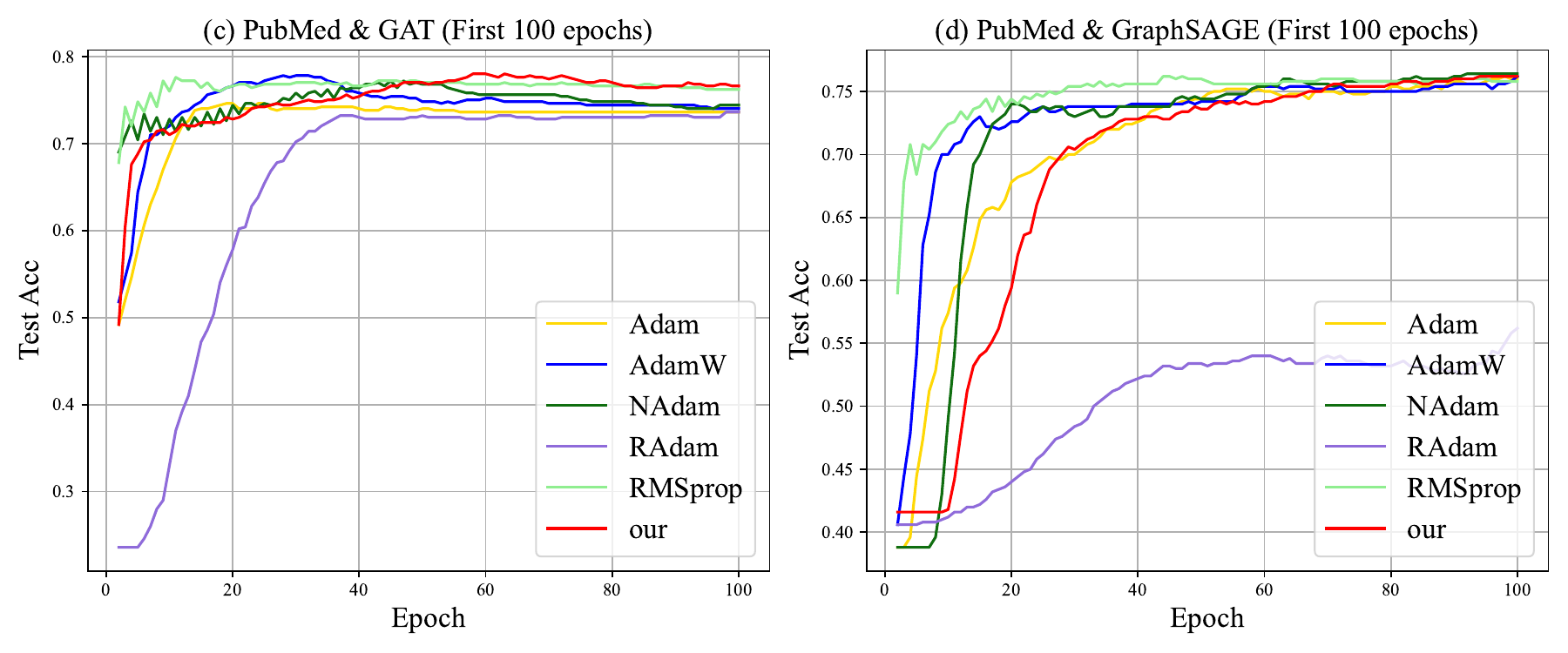}
	\end{subfigure}
	\caption{Comparison of test accuracy for the Cora and PubMed datasets on the GAT and GraphSAGE benchmark model. (a) GAT testing on Cora dataset. (b) GraphSAGE testing on Cora. (c) GAT testing on PubMed dataset. (d) GraphSAGE testing on PubMed dataset.}
	\label{f:11}
\end{figure}

As shown in Fig. \ref{f:9}, our method achieves stable and superior performance across both Accuracy and F1 Score in the Core dataset, maintaining an advantage over Adam, RMSprop, and AdamW. Similarly, Fig. \ref{f:10} highlights the performance gains in the PubMed dataset, where our method consistently leads over other optimizers. These results indicate that our proposed approach provides a more robust and generalizable optimization strategy across different datasets and graph learning architectures. Additionally, we visualize the optimization process of the baseline models, as shown in the Fig. \ref{f:11}.

\subsection{Audio Classification}

In the field of speech processing, which is one of the three classic tasks along with computer vision and natural language processing in the field of artificial intelligence, speech tasks constitute an important domain within deep learning, with mature models and datasets. In this section, we conduct experiments on audio classification using the UrbanSound8K \cite{rahman2021classification} dataset. The baseline models we selected include CAMP++ \cite{wang2023cam}, TDNN \cite{Desplanques_2020}, and Res2net \cite{gao2019res2net}. In Table \ref{t:9}, we report the experimental results for each optimizer. Additionally, we provide the time consumption of the models during the evaluation phase to assess the impact of our optimizer on time overhead.

On the CAM++ and TDNN models, our optimizer achieved classification accuracies comparable to or even higher than SGD and NAdam, surpassing Adam, AdamW, and others. Meanwhile, on the Res2net model, our optimizer achieved the best classification results. This aligns with the conclusions drawn from the earlier experiments in image classification and natural language processing, demonstrating the task-independence of our optimizer.

The results, as presented in Table \ref{t:9}, compare the evaluation times of our optimizer against SGD, Adam, AdamW, RAdam, NAdam, and RMSprop. The performance of our method across three models (CAM++, TDNN, Res2net) closely matches or surpasses that of other optimizers. Specifically, it achieved an accuracy of 0.94877 on the CAM++ model, narrowly trailing SGD's 0.95079; it registered 0.92276 on the TDNN model, nearly reaching NAdam's peak score of 0.92347; and it attained the highest accuracy of 0.90256 on the Res2net model, outperforming all other optimizers. This consistently strong performance indicates that our method likely possesses robust generalization capabilities, making it effective across different model architectures beyond just a single model.

\begin{table}[htbp]
	\centering
	\caption{Accuracy on the audio categorization task with 95\% confidence intervals.}
	\label{t:9}
	\begin{tabular}{p{0.2\linewidth}ccc}
		\hline
		\textbf{Model} & \textbf{CAM++ (\%)} & \textbf{TDNN (\%)} & \textbf{Res2net (\%)} \\ \hline
		SGD & \textbf{95.08 ± 0.03} & 86.32 ± 0.02 & 73.57 ± 0.04 \\
		AdamW & 90.90 ± 0.04 & 90.72 ± 0.03 & 89.01 ± 0.02 \\
		Adam & 90.45 ± 0.02 & 92.12 ± 0.04 & 90.19 ± 0.01 \\
		RAdam & 92.07 ± 0.03 & 91.84 ± 0.02 & 86.73 ± 0.04 \\
		NAdam & 92.29 ± 0.02 & \textbf{92.35 ± 0.03} & 85.06 ± 0.01 \\
		RMSprop & 94.41 ± 0.04 & 89.11 ± 0.02 & 85.66 ± 0.03 \\
		our & 94.88 ± 0.01 & 92.28 ± 0.04 & \textbf{90.26 ± 0.02} \\ \hline
	\end{tabular}
\end{table}

\subsection{Experimental Evaluation of the Convergence}

The Rosenbrock function \cite{shang2006note}, introduced by Howard H. Rosenbrock in 1960, is a renowned benchmark in unconstrained optimization due to its unique non-convex shape, often referred to as Rosenbrock's valley or the Rosenbrock banana function. We utilize solving it to observe the trajectory change and convergence rate of DWMGrad. The Rosenbrock function is defined as:
\begin{equation}
	f(x,y)=(1-x)^2+100(y-x^2)^2
\end{equation}
its contours resemble a series of parabolic shapes, with the global minimum at $(x,y)=(1,1)$ where $f(x,y)=0$. Despite the ease of finding the valley, precisely locating the global minimum poses a significant challenge due to subtle changes in function value within the valley. This makes the Rosenbrock function invaluable for evaluating optimization algorithms, showcasing their behavior in narrow and curved search spaces.

Our focus is on the two-dimensional case, comparing our optimization algorithm with Adam and AdaGrad. Through this comparison, we aim to demonstrate our algorithm's efficiency and stability in finding global optima, particularly in navigating tortuous paths.

\begin{figure}[ht]
	\centering
	\begin{subfigure}[b]{0.32\linewidth}
		\includegraphics[width=\linewidth]{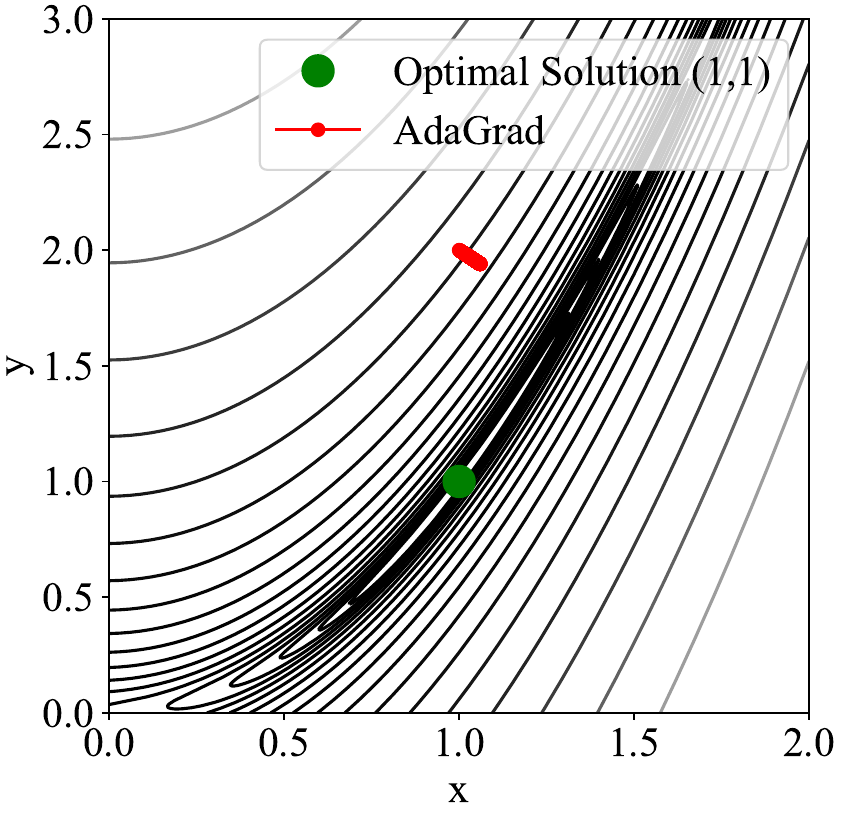}
		\caption{AdaGrad optimizer.} 
	\end{subfigure}
	\hspace{0.cm}
	\begin{subfigure}[b]{0.32\linewidth}
		\includegraphics[width=\linewidth]{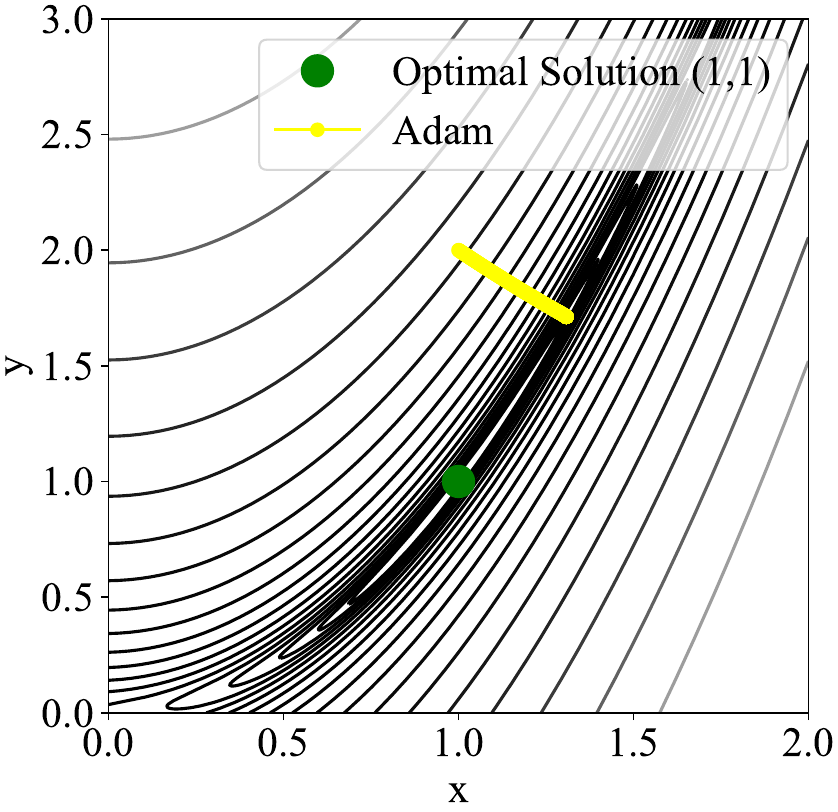}
		\caption{Adam optimizer.}
	\end{subfigure}
	\hspace{0.cm}
	\begin{subfigure}[b]{0.32\linewidth}
		\includegraphics[width=\linewidth]{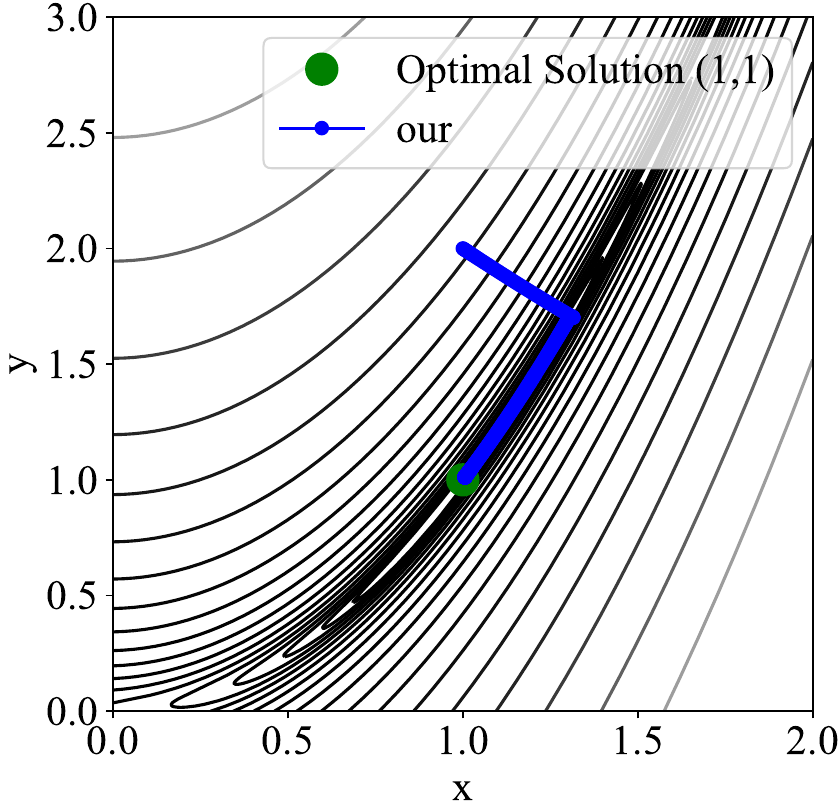}
		\caption{Our optimizer.}
	\end{subfigure}
	\caption{Path change of the Rosenbrock function when the minimum is $(1,1)$ for 1000 iterations of optimization using AdaGrad, Adam and our method. The green point indicates the optimal point, the red path is the optimization path of AdaGrad, the yellow path is the optimization path of Adam, and the blue path is the optimization path of our method.}
	\label{f:12}
\end{figure}

Fig. \ref{f:12} illustrates the trajectory tracking of AdaGrad, Adam, and our method over 1000 iterations on the Rosenbrock function. Our method approaches the optimal solution more closely compared to others.

\begin{figure}[htbp]
	\centering
		\begin{subfigure}[b]{0.48\linewidth}
		\includegraphics[width=\linewidth]{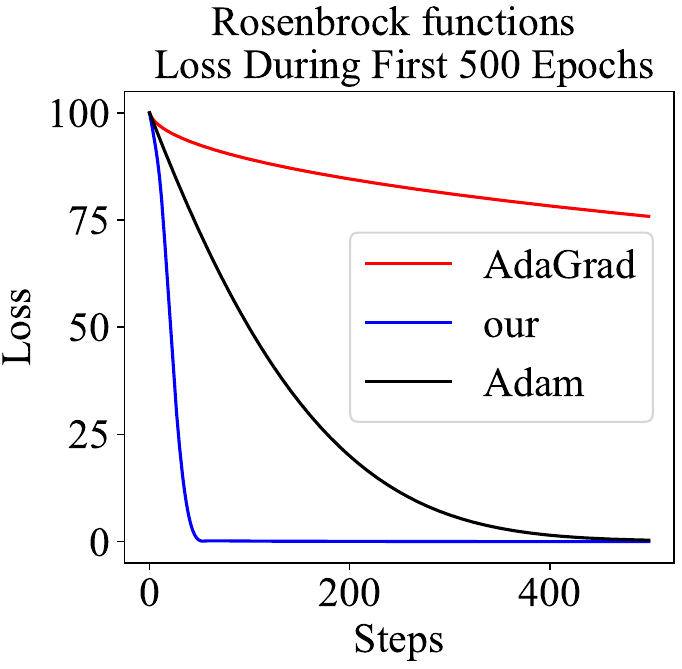}
		\caption{Comparions on Rosenbrock function.} 
	\end{subfigure}
	\hspace{0.cm}
	\begin{subfigure}[b]{0.49\linewidth}
		\includegraphics[width=\linewidth]{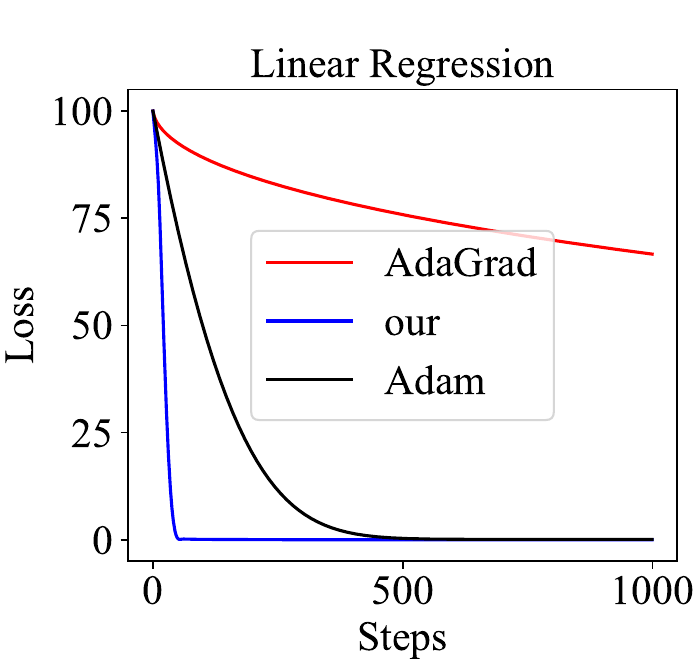}
		\caption{Comparions on linear regression.}
	\end{subfigure}
	\caption{Losses when optimizing the Rosenbrock function using methods using AdaGrad, Adam, and ours. The left panel shows the loss for the first 500 epochs and the right panel shows the complete loss.}
	\label{f:13}
\end{figure}

Fig. \ref{f:13} displays the loss value evolution for each optimizer in the search for the Rosenbrock function's optimal solution. Our method exhibits a faster rate of decrease, indicating swift progress towards the optimum. It maintains a steady downward trend, suggesting stable progression towards the optimal solution, which is advantageous for solving complex problems.

\subsection{Experimental Evaluation of the Computational Complexity}
\label{sec4.5}

We use the node classification task as an example and monitor the actual runtime in Table \ref{t:10} to validate our analysis and provide a practical perspective on computational requirements.

\begin{table}[htbp]
	\centering
	\caption{Time cost on Cora and PubMed datasets. Reported values are mean ± standard deviation over multiple runs.}
	\label{t:10}
	\begin{tabular}{lcc|cc}
		\hline
		\multirow{2}{*}{\textbf{Task}} & \multicolumn{2}{c}{\textbf{Cora}} & \multicolumn{2}{c}{\textbf{Pubmed}} \\ \cline{2-5}
		& GAT & GraphSAGE & GAT & GraphSAGE \\ \hline
		Adam & 160s ± 3s & \textbf{120s ± 2s} & 120s ± 4s & 200s ± 3s \\
		AdamW & 165s ± 4s & 160s ± 3s & \textbf{100s ± 2s} & 195s ± 4s \\
		\textbf{our} & \textbf{30s ± 2s} & 160s ± 4s & 120s ± 3s & \textbf{200s ± 2s} \\ \hline
	\end{tabular}
\end{table}

The results show that our method exhibits obvious advantages in the node classification task, especially on the GAT task of Cora dataset. This suggests that our method may be well tuned for a specific architecture and dataset and able to greatly accelerate model training.

\section{Conclusion}
\label{sec:5}
In this paper, we proposed a novel optimizer based on a dynamic sliding window mechanism that adjusts the range of historical information dynamically during optimization. By employing a sliding window approach, our optimizer can flexibly adapt the span of historical data to suit various training scenarios. This dynamic adjustment mechanism enables the optimizer to better handle changes and complexities in different tasks, leading to enhanced performance across diverse situations. The proposed DWMGrad Optimizer, built upon the adaptive momentum optimizer Adam, demonstrates its effectiveness and superiority through rigorous theoretical analysis and a series of empirical experiments.

Our experimental results demonstrate that the DWMGrad optimizer consistently delivers superior performance across a wide range of tasks and domains. Specifically, in image classification tasks, it achieves higher final accuracy on small, medium, and large-scale datasets. The optimizer also achieves state-of-the-art performance on multiple tasks in the GLUE benchmark, particularly excelling in graph node classification tasks where it not only achieves higher final accuracy but also demonstrates the fastest convergence rate. Furthermore, DWMGrad maintains competitive performance with current optimizers in speech classification tasks. Additionally, we have validated the algorithm's universality and robustness through extensive testing on the Rosenbrock function.

Despite its advantages, the DWMGrad optimizer has certain limitations. The theoretical convergence analysis presented in this work is conducted under the convex optimization assumption, which is commonly adopted in optimization literature to provide analytical tractability. However, we acknowledge that this assumption does not fully capture the inherently non-convex nature of most deep learning problems. As such, the theoretical results may not be directly applicable to all practical scenarios. We have clarified this limitation explicitly in the conclusion to avoid any misinterpretation of the theoretical scope.

To address this gap, future work will focus on extending the theoretical analysis of DWMGrad under non-convex settings, aligning more closely with real-world deep learning applications. Additionally, we plan to evaluate the optimizer on more complex and large-scale tasks, including reinforcement learning, large language models, and multimodal learning scenarios. We also aim to improve the computational efficiency of DWMGrad to support large-scale deployment. These efforts will contribute to a more comprehensive understanding of the optimizer's behavior and its practical utility across a broader range of applications.

\section*{Declarations}
\subsection*{Conflict of interest} The authors declare that they have no conflict of interest.

\subsection*{Data availability}
Data will be made available on request.

\bibliography{ref,Citations}
%


\end{document}